\documentclass{article}

\PassOptionsToPackage{numbers, compress}{natbib}
\usepackage[ruled]{algorithm2e}
\usepackage{algpseudocode}
\usepackage[final]{neurips_2019}

\usepackage[utf8]{inputenc} 
\usepackage[T1]{fontenc}    
\usepackage{hyperref}       
\usepackage{url}            
\usepackage{booktabs}       
\usepackage{amsfonts}       
\usepackage{nicefrac}       
\usepackage{microtype}      
\usepackage{tabularx}
\usepackage{multirow}

\usepackage{mathtools}
\usepackage{amsthm}
\usepackage{amssymb}
\usepackage{amsmath}
\usepackage{tablefootnote}
\usepackage{thm-restate}
\usepackage[separate-uncertainty=true]{siunitx}
\usepackage{longtable}

\DeclareMathOperator*{\argmin}{arg\,min}
\DeclareMathOperator{\diag}{diag}
\DeclareMathOperator{\rank}{rank}
\DeclareMathOperator{\spn}{span}
\DeclareMathOperator{\Tr}{Tr}

\usepackage{subfiles}

\newtheorem{theorem}{Theorem}

\newtheorem{remark}{Remark}[theorem]

\newtheorem{lemma}{Lemma}[theorem]

\newtheorem{lemmasec}{Lemma}[section]
\newtheorem{corollary}{Corollary}[lemma]

\newtheorem{definition}{Definition}
\newtheorem{remarkdef}{Remark}[definition]

\usepackage{cancel}
\usepackage{xcolor}
\usepackage{bm}
\newcommand{\bx}{\mathbf{x}}
\newcommand{\by}{\mathbf{y}}

\DeclareMathOperator*{\Gr}{Gr}
\DeclareMathOperator*{\Lip}{Lip}

\DeclareMathOperator*{\argmax}{arg\,max}
\title{Preventing Gradient Attenuation in \\ Lipschitz Constrained Convolutional Networks}

\author{
   Qiyang Li$^*$, Saminul Haque\thanks{Equal contributions}~ , Cem Anil, James Lucas, Roger Grosse, J\"orn-Henrik Jacobsen \\
   University of Toronto, Vector Institute \\
   \texttt{\{qiyang.li, saminul.haque, cem.anil\}@mail.utoronto.ca} \\
   \texttt{\{jlucas, rgrosse\}@cs.toronto.edu} \\
   \texttt{j.jacobsen@vectorinstitute.ai}
}

\begin{document}
\maketitle

\begin{abstract}
    Lipschitz constraints under $L_2$ norm on deep neural networks are useful for provable adversarial robustness bounds, stable training, and Wasserstein distance estimation. While heuristic approaches such as the gradient penalty have seen much practical success, it is challenging to achieve similar practical performance while provably enforcing a Lipschitz constraint. In principle, one can design Lipschitz constrained architectures using the composition property of Lipschitz functions, but \citet{anil2018sorting} recently identified a key obstacle to this approach: gradient norm attenuation. They showed how to circumvent this problem in the case of fully connected networks by designing each layer to be gradient norm preserving. We extend their approach to train scalable, expressive, provably Lipschitz convolutional networks. In particular, we present the Block Convolution Orthogonal Parameterization (BCOP), an expressive parameterization of orthogonal convolution operations. We show that even though the space of orthogonal convolutions is disconnected, the largest connected component of BCOP with $2n$ channels can represent arbitrary BCOP convolutions over $n$ channels. Our BCOP parameterization allows us to train large convolutional networks with provable Lipschitz bounds. Empirically, we find that it is competitive with existing approaches to provable adversarial robustness and Wasserstein distance estimation.~\footnote{Code is available at: \url{github.com/ColinQiyangLi/LConvNet}}
\end{abstract}

\section{Introduction}
There has been much interest in training neural networks with known upper bounds on their Lipschitz constants under $L_2$ norm\footnote{Unless specified otherwise, we refer to Lipschitz constant as the Lipschitz constant under $L_2$ norm.}. Enforcing Lipschitz constraints can provide provable robustness against adversarial examples~\citep{tsuzuku2018lipschitz}, improve generalization bounds~\citep{sokolic2017robust}, and enable Wasserstein distance estimation~\citep{anil2018sorting, arjovsky2017wasserstein, gulrajani2017improved}. Heuristic methods for enforcing Lipschitz constraints, such as the gradient penalty~\citep{gulrajani2017improved} and spectral norm regularization~\citep{yoshida2017spectral}, have seen much practical success, but provide no guarantees about the Lipschitz constant. It remains challenging to achieve similar practical success while provably satisfying a Lipschitz constraint.

In principle, one can design provably Lipschitz-constrained architectures by imposing a Lipschitz constraint on each layer; the Lipschitz bound for the network is then the product of the bounds for each layer. \citet{anil2018sorting} identified a key difficulty with this approach: because a layer with a Lipschitz bound of 1 can only reduce the norm of the gradient during backpropagation, each step of backprop gradually attenuates the gradient norm, resulting in a much smaller Jacobian for the network's function than is theoretically allowed. We refer to this problem as \emph{gradient norm attenuation}. They showed that Lipschitz-constrained ReLU networks were prevented from using their full nonlinear capacity due to the need to prevent gradient norm attenuation. To counteract this problem, they introduced \emph{gradient norm preserving (GNP)} architectures, where each layer preserves the gradient norm. For fully connected layers, this involved constraining the weight matrix to be orthogonal and using a GNP activation function called GroupSort. Unfortunately, the approach of \citet{anil2018sorting} only applies to fully-connected networks, leaving open the question of how to constrain the Lipschitz constants of convolutional networks.

As many state-of-the-art deep learning applications rely on convolutional networks, there have been numerous attempts for tightly enforcing Lipschitz constants of convolutional networks. However, those existing techniques either hinder representational power or induce difficulty in optimization. \citet{cisse2017parseval, tsuzuku2018lipschitz, qian2018l2} provide loose bounds on the Lipschitz constant that can limit the parameterizable region. \citet{gouk2018regularisation} obtain a tight bound on the Lipschitz constant, but tend to lose expressive power during training due to vanishing singular values. The approach of \citet{sedghi2018singular} is computationally intractable for larger networks. 

In this work, we introduce convolutional GNP networks with an efficient parameterization of orthogonal convolutions by adapting the construction algorithm from \citet{xiao2018dynamical}. This parameterization avoids the issues of loose bounds on the Lipschitz constant and computational intractability observed in the aforementioned approaches. Furthermore, we provide theoretical analysis that demonstrates the disconnectedness of the orthogonal convolution space, and how our parameterization alleviates the optimization challenge engendered by the disconnectedness. 

We evaluate our GNP networks in two situations where expressive Lipschitz-constrained networks are of central importance. The first is provable norm-bounded adversarial robustness, which is the task of classification and additionally certifying that the network's classification will not change under any norm-bounded perturbation. Due to the tight Lipschitz properties, the constructed GNP networks can easily give non-trivial lower bounds on the robustness of the network's classification. We demonstrate that our method outperforms the state-of-the-art in provable deterministic robustness under $L_2$ metric on MNIST and CIFAR10. The other application is Wasserstein distance estimation. Wasserstein distance estimation can be rephrased as a maximization over 1-Lipschitz functions, allowing our Lipschitz-constrained networks to be directly applied to this problem. Moreover, the restriction to GNP we impose is not necessarily a hindrance, as it is shown by \citet{gemici2018primal} that the optimal 1-Lipschitz function is also GNP almost everywhere. We demonstrate that our GNP convolutional networks can obtain tighter Wasserstein distance estimates than competing architectures.

\section{Background}
\subsection{Lipschitz Functions under \texorpdfstring{$L_2$}{L2} Norm}
In this work, we focus on Lipschitz functions with respect to the $L_2$ norm. We say a function $f: \mathbb{R}^n \rightarrow \mathbb{R}^m$ is $l$-Lipschitz if and only if
\begin{equation}
    ||f(\bx_1) - f(\bx_2)||_2 \leq l  ||\bx_1 - \bx_2||_2, \forall \bx_1, \bx_2 \in \mathbb{R}^n
\end{equation}
We denote $\Lip(f)$ as the smallest $K$ for which $f$ is $l$-Lipschitz, and call it the \textbf{Lipschitz constant} of $f$. 
For two Lipschitz continuous functions $f$ and $g$, the following property holds:
\begin{align}
\label{eq:lip-comp}
    \Lip(f \circ g) \leq \Lip(f)\Lip(g) 
\end{align}
The most basic neural network design consists of a composition of linear transformations and non-linear activation functions. The property above (Equation \ref{eq:lip-comp}) allows one to upper-bound the Lipschitz constant of a network by the product of the Lipschitz constants of each layer. However, as modern neural networks tend to possess many layers, the resultant upper-bound is likely to be very loose, and constraining it increases the risk of diminishing the Lipschitz constrained network capacity that can be utilized.

\subsection{Gradient Norm Preservation (GNP)}
\label{sec:gnp}
Let $\by=f(\bx)$ be 1-Lipschitz, and $\mathcal{L}$ be a loss function. The norm of the gradient after backpropagating through a 1-Lipschitz function is no larger than the norm of the gradient before doing so:
\begin{align*}
    \left\|\nabla_{\bx} \mathcal{L} \right\|_2 = \left\|(\nabla_{\by} \mathcal{L}) (\nabla_{\bx} f) \right\|_2 \leq  \left\|\nabla_{\by} \mathcal{L} \right\|_2 \left\|\nabla_{\bx} f \right\|_2 \leq \Lip(f) \left\|\nabla_{\by} \mathcal{L}\right\|_2  \leq \left\|\nabla_{\by} \mathcal{L}\right\|_2
\end{align*}

As a consequence of this relation, the gradient norm will likely be attenuated during backprop if no special measures are taken. One way to fix the gradient norm attenuation problem is to enforce each layer to be gradient norm preserving (GNP). Formally, $f: \mathbb{R}^n \mapsto \mathbb{R}^m$ is GNP if and only if its input-output Jacobian, $J\in\mathbb{R}^{m\times n}$, satisfies the following property:
\begin{align*}
     \left\|J^T\mathbf{g}\right\|_2 = ||\mathbf{g}||_2, \forall \mathbf{g} \in \mathcal{G}.
\end{align*}
where $\mathcal{G} \subseteq \mathbb{R}^{m}$ defines the possible values that the gradient vector $\mathbf{g}$ can take. Note that when $m=n$, this condition is equivalent to orthogonality of $J$. In this work, we consider a slightly stricter definition where $\mathcal{G} = \mathbb{R}^{m}$ because this allows us to directly compose two GNP (strict) functions without reasoning about their corresponding $\mathcal{G}$. For the rest of the paper, unless specified otherwise, a GNP function refers to this more strict definition.

Based on the definition of GNP, we can deduce that GNP functions are 1-Lipschitz in the 2-norm. Since the composition of GNP functions is also GNP, one can design a GNP network by stacking GNP building blocks. Another favourable condition that GNP networks exhibit is \emph{dynamical isometry}~\citep{xiao2018dynamical, pennington2017resurrecting, pmlr-v84-pennington18a} (where the entire distribution of singular values of input-output Jacobian is close to 1), which has been shown to improve training speed and stability.

\subsection{Provable Norm-bounded Adversarial Robustness}
\label{sec:background-adv}
We consider a classifier $f$ with $T$ classes that takes in input $\bx$ and produces a logit for each of the classes: $f(\bx) = \begin{bmatrix} y_1 & y_2 & \cdots & y_{T} \end{bmatrix}$. An input data point $\bx$ with label $t \in \{1, 2, \cdots, T\}$ is provably robustly classified by $f$ under perturbation norm of $\epsilon$ if
\begin{align*}
    \argmax_i f(\bx + \bm{\delta})_i = t, \forall \bm{\delta}: ||\bm{\delta}||_2 \leq \epsilon.
\end{align*}
The margin of the prediction for $\bx$ is given by $\mathcal{M}_f(\bx) = \max(0, y_t - \max_{i\neq t} y_i)$.
If $f$ is $l$-Lipschitz, we can certify that $f$ is robust with respect to $\bx$ if $\sqrt{2}l\epsilon < \mathcal{M}_f(\mathbf{x})$ (See Appendix \ref{sec:appendix-adv-robust} for the proof). 

\subsection{Wasserstein Distance Estimation}
Wasserstein distance is a distance metric between two probability distributions \citep{peyre2018computational}. The Kantorovich-Rubinstein formulation of Wasserstein distance expresses it as a maximization problem over 1-Lipschitz functions \citep{arjovsky2017wasserstein}: 
\begin{equation}
\label{eq:dual_obj}
    W(P_{1}, P_{2}) = \sup_{f : \Lip(f) \leq 1} \left(  \mathop{\mathbb{E}}_{\bx \sim P_{1}(\bx)}[f(\bx)] -  \mathop{\mathbb{E}}_{\bx \sim P_{2}(\bx)}[f(\bx)]  \right).
\end{equation}
In Wasserstein GAN architecture, \citet{arjovsky2017wasserstein} proposed to parametrize the scalar-valued function $f$ using a Lipschitz constrained network, which serves as the discriminator that estimates the Wasserstein distance between the generator and data distribution. One important property to note is that the optimal scalar function $f$ is GNP almost everywhere (See Corollary 1 in \citet{gemici2018primal}). Naturally, this property favours the optimization approach that focuses on searching over GNP functions. Indeed, \citet{anil2018sorting} found that GNP networks can achieve tighter lower bounds compared to non-GNP networks.

\section{Orthogonal Convolution Kernels}
The most crucial step to building a GNP convolutional network is constructing the GNP convolution itself. Since a convolution operator is a linear operator, making the convolution kernel GNP is equivalent to making its corresponding linear operator orthogonal. While there are numerous methods for orthogonalizing arbitrary linear operators, it is not immediately clear how to do this for convolutions, especially when preserving kernel size. We first summarize the orthogonal convolution representations from \citet{kautsky1994matrix} and \citet{xiao2018dynamical} (Section \ref{sec:construct-orthogonal-conv}). Then, we analyze the topology of the space of orthogonal convolution kernels and demonstrate that the space is disconnected (with at least $\mathcal{O}(n^2)$ connected components for a $2\times 2$ 2-D convolution layer), which is problematic for gradient-based optimization methods because they are confined to one component (Section \ref{sec:intriguing}). Fortunately, this problem can be fixed by increasing the number of channels: we demonstrate that a single connected component of the space of orthogonal convolutions with $2n$ channels can represent any orthogonal convolution with $n$ channels (Section \ref{sec:lip-conv-bcop}).

\subsection{Constructing Orthogonal Convolutions}
\label{sec:construct-orthogonal-conv}
To begin analysing orthogonal convolution kernels, we must first understand the symmetric projector, which is a fundamental building block of orthogonal convolutions. An $n\times n$ matrix $P$ is defined to be a symmetric projector if $P = P^2 = P^T$. Geometrically, a symmetric projector $P$ represents an orthogonal projection onto the range of $P$. From this geometric interpretation, it is not hard to see that the space of projectors has $n+1$ connected components, based on the rank of the projector (for a more rigorous treatment, see Remark \ref{remark:sym} in Appendix \ref{appendix:disconnectedness}). For notation simplicity, we denote $\mathbb{P}(n)$ as the set of all $n \times n$ symmetric projectors and $\mathbb{P}(n, k)$ as the subset of all $n \times n$ symmetric projectors with ranks of $k$.

\begin{figure}
    \centering
    \includegraphics[width=\linewidth]{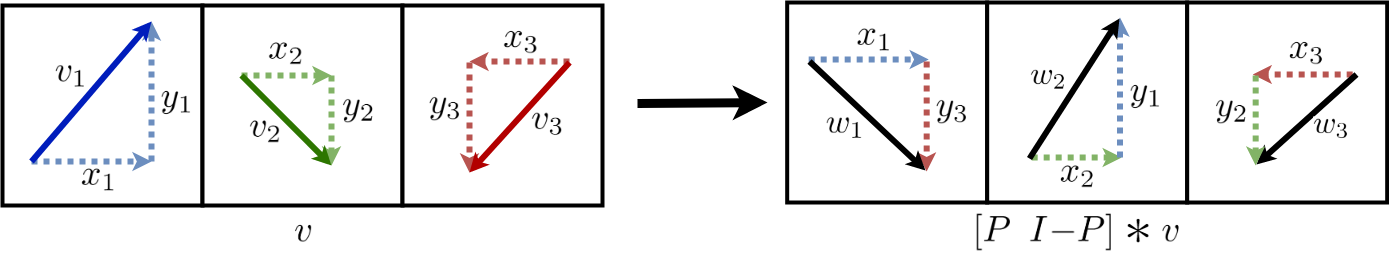}
    \caption{Visualization of a 1-D orthogonal convolution, $[P\ \ I-P]$, applied to a 1-D input tensor $v \in \mathbb{R}^{2\times 3}$ with a length of 3 and channel size of 2. $P \in \mathbb{R}^{2\times 2}$ here is the orthogonal projection onto the $x$-axis, which makes $I-P$ the complementary projection onto the $y$-axis. Each cell of $v$ corresponds to one of the three spatial locations, and the vector contained within it represents the vector along the channel dimension in said spatial location.}
    \label{fig:bconv-viz}
\end{figure}

Now that the concept of symmetric projectors has been established, we will consider how to construct 1-D convolutional kernels. As shown by \citet{kautsky1994matrix}, all 1-D orthogonal convolution kernels with a kernel size $K$ can be represented as:
\begin{equation}
\begin{aligned}
    \mathcal{W}(H, P_{1:K-1}) = H \square \begin{bmatrix}P_1 & (I-P_1)\end{bmatrix} \square \cdots \square \begin{bmatrix}P_{K-1} & (I-P_{K-1})\end{bmatrix}
    \label{eq:1d-param}
\end{aligned}
\end{equation}
where $H \in O(n)$ is an orthogonal matrix of $n\times n$, $P_i \in \mathbb{P}(n)$, and $\square$ represents block convolution, which is convolution using matrix multiplication rather than scalar multiplication:
\begin{equation*}
    \begin{bmatrix}X_1 & X_2 & \cdots & X_p \end{bmatrix} \square \begin{bmatrix} Y_1 & Y_2 & \cdots & Y_q \end{bmatrix} = \begin{bmatrix}Z_1 & Z_2 & \cdots & Z_{p+q-1} \end{bmatrix}
\end{equation*} with $Z_i = \sum_{i'=-\infty}^{\infty} X_{i'} Y_{i-i'} $, where the out-of-range elements are all zero (e.g., $X_{<1} = 0, X_{>p} = 0, Y_{<1} = 0, Y_{>q} = 0$). Unlike regular convolutions, the block convolution does not commute since matrix multiplication does not commute.
One important property of block convolution is that it corresponds to composition of the kernel operators. That is, $X \ast \left(Y \ast v \right) = \left(X \square Y\right) \ast v$, where $A \ast v$ represents the resulting tensor after applying convolution $A$ to $v$. This composition property allows us to decompose the representation (Equation \ref{eq:1d-param}) into applications of orthogonal convolutions with kernel size of 2 (Figure \ref{fig:bconv-viz} demonstrates the effect of it) along with a channel-wise orthogonal transformation ($H$).

\citet{xiao2018dynamical} extended the 1-D representation to the 2-D case using alternating applications of orthogonal convolutions of size 2:
\begin{equation}
\begin{aligned}
    \mathcal{W}(H, P_{1:K-1}, Q_{1:K-1}) = H \square& \begin{bmatrix} P_1 \\ I-P_1 \end{bmatrix} \square \begin{bmatrix}Q_1 & I-Q_1 \end{bmatrix} \square \cdots \\ \cdots&\square \begin{bmatrix} P_{K-1} \\ I-P_{K-1} \end{bmatrix}\square \begin{bmatrix} Q_{K-1} & I-Q_{K-1} \end{bmatrix}
    \label{eq:2d-param}
\end{aligned}
\end{equation}
where $Z = X\square Y$ is defined similarly to the 1-D case with $Z_{ij} = \sum_{i'=-\infty}^{\infty}\sum_{j'=-\infty}^{\infty} \left[X_{i', j'} Y_{i-i', j-j'}\right]$, and $P_i, Q_i \in \mathbb{P}(n)$. Unlike in 1-D, we discovered that this 2-D representation could only represent a subset of 2-D orthogonal convolutions (see Appendix \ref{appendix:2d-incompleteness} for an example). However, we do not know whether simple modifications to this parameterization will result in a complete representation of all 2-D orthogonal convolutions (see Appendix \ref{appendix:2d-incompleteness} for details on the open question). 

\subsection{Topology of the Orthogonal Convolution Space}
\label{sec:intriguing}
Before utilizing this space of orthogonal convolutions, we would like to analyze some fundamental properties of this space. Since $\mathbb{P}(n)$ has $n+1$ connected components and orthogonal convolutions are constructed out of many projectors, it is to be expected that there are many connected components in the space of orthogonal convolutions. Indeed, we see the first result in 1-D (Theorem \ref{thm:1dbcop}). 

\begin{restatable}[Connected Components of 1-D Orthogonal Convolution]{theorem}{onedbcop}
The 1-D orthogonal convolution space is compact and has $2(K-1)n+2$ connected components, where $K$ is the kernel size and $n$ is the number of channels.
\label{thm:1dbcop}
\end{restatable}

In 2-D, we analyze case of kernel size of $2$ ($2\times 2$ kernels) and show that the number of connected components grows at least quadratically with respect to the channel size:

\begin{restatable}[Connected Components of 2-D Orthogonal Convolution with $K=2$]{theorem}{twodbcop}
2-D orthogonal convolution space with a kernel size of $2\times 2$ has at least $2(n+1)^2$ connected components, where $n$ is the number of channels.
\label{thm:2dbcop}
\end{restatable}

The disconnectedness in the space of orthogonal convolution imposes an intrinsic difficulty in optimizing over the space of orthogonal convolution kernels, as gradient-based optimizers are confined to their initial connected component. We refer readers to Appendix \ref{appendix:disconnectedness} for the proof of Theorem \ref{thm:1dbcop} and Appendix \ref{appendix:2d-disconnectedness} for the proof of Theorem \ref{thm:2dbcop}.

\begin{algorithm}[t]
\SetAlgoLined
\KwIn{$c_o \times c_i$ unconstrained matrix $O$, $c_i \times \left\lfloor\frac{c_i}{2}\right\rfloor$ unconstrained matrices $M_i,N_i$ for $i$ from 1 to $k-1$, assuming $c_i \geq c_o$}
\KwResult{Orthogonal Convolution Kernel $W \in \mathbb{R}^{c_o \times c_i \times K \times K}$ }
 $H \leftarrow \mathrm{Orthogonalize}(O)$; \Comment{any differentiable orthogonalization procedure (e.g., Bj\"orck~\citep{bjorck1971iterative})}\;
 Initialize $W$ as a $1\times 1$ convolution with $W[0,0]=H$\;
 \For{$i$ from 1 to $K-1$}{
  $R_P,\ R_Q \leftarrow \mathrm{Orthogonalize}(M_i),\  \mathrm{Orthogonalize}(N_i)$\;
  $P,\ Q \leftarrow R_PR_P^T,\ R_QR_Q^T$; \Comment{Construct symmetric projectors with half of the full rank}\;
  $W \leftarrow W\square\begin{bmatrix} P \\ I-P \end{bmatrix} \square \begin{bmatrix}Q & I-Q \end{bmatrix}$
 }
 \KwOut{$W$}
\label{algo:jeff}
 \caption{Block Convolution Orthogonal Parameterization (BCOP)}
\end{algorithm}

\subsection{Block Convolution Orthogonal Parameterization (BCOP)}
\label{sec:lip-conv-bcop}

To remedy the disconnectedness issue, we show the following:
\begin{restatable}[BCOP Construction with Auxiliary Dimension]{theorem}{bcopdouble}
    For any convolution $C=\mathcal{W}(H, P_{1:K-1}, Q_{1:K-1})$ with input and output channels $n$ and $P_i, Q_i \in \mathbb{P}(n)$, there exists a convolution $C'=\mathcal{W}(H', P'_{1:K-1}, Q'_{1:K-1})$ with input and output channels $2n$ constructed from only $n$-rank projectors ($P'_i, Q'_i \in \mathbb{P}(2n, n)$) such that $C'(\mathbf{x})_{1:n} = C(\mathbf{x}_{1:n})$. That is, the first $n$ channels of the output is the same with respect to the first $n$ channels of the input under both convolutions.
    \label{thm:bcop-aux}
\end{restatable}

The idea behind this result is that some projectors in $\mathbb{P}(2n,n)$ may use their first $n$ dimensions to represent $\mathbb{P}(n)$ and then use the latter $n$ dimensions in a trivial capacity so that the total rank is $n$ (see Appendix \ref{sec:double} for the detailed proof).

Theorem~\ref{thm:bcop-aux} implies that all connected components of orthogonal convolutions constructed by $\mathcal{W}$ with $n$ channels can all be equivalently represented in a single connected component of convolutions constructed by $\mathcal{W}$ with $2n$ channels by only using projectors that have rank $n$. (This comes at the cost of requiring 4 times as many parameters.)

This result motivates us to parameterize the connected subspace of orthogonal convolutions defined by $\mathcal{W}(H, \Tilde{P}_{1:K-1}, \Tilde{Q}_{1:K-1})$ where $\Tilde{P}_i \in \mathbb{P}(n,  \left\lfloor\frac{n}{2}\right\rfloor)$ and $\Tilde{Q}_i \in \mathbb{P}(n, \left\lfloor\frac{n}{2}\right\rfloor)$. We refer to this method as the Block Convolution Orthogonal Parameterization (BCOP). The procedure for BCOP is summarized in Algorithm \ref{algo:jeff} (See Appendix \ref{appendix:bcop-procedure} for implementation details).

\section{Related Work}
\label{sec:lip-conv-exist}
\paragraph{Reshaped Kernel Method (RK)}
This method reshapes a convolution kernel with dimensions $(c_o,c_i,k,k)$ into a $(c_o, k^2c_i)$ matrix. The Lipschitz constant (or spectral norm) of a convolution operator is bounded by a constant factor of the spectral norm of its reshaped matrix~\citep{cisse2017parseval, tsuzuku2018lipschitz, qian2018l2}, which enables bounding of the convolution operator's Lipschitz constant by bounding that of the reshaped matrix. However, this upper-bound can be conservative, causing a bias towards convolution operators with low Lipschitz constants, limiting the method's expressive power. In this work, we strictly enforce orthogonality of the reshaped matrix rather than softly constrain it via regularization, as done in \citet{cisse2017parseval}. We refer to this variant as \textbf{reshaped kernel orthogonalization}, or \textbf{RKO}.

\paragraph{One-Sided Spectral Normalization (OSSN)} 
This variant of spectral normalization~\citep{miyato2018spectral} scales the kernel so that the spectral norm of the convolution operator is at most 1~\citep{gouk2018regularisation}.
This is a projection under the matrix 2-norm but \emph{not} the Frobenius norm. It is notable because when using Euclidean steepest descent with this projection, such as in constrained gradient-based optimization, there is no guarantee to converge to the correct solution (see an explicit example and further analysis in Appendix~\ref{app:optimizing_sn}). In practice, we found that projecting during the forward pass (as in \citet{miyato2018spectral}) yields better performance than projecting after each gradient update.

\paragraph{Singular Value Clipping and Masking (SVCM)}
Unlike spectral normalization, singular value clipping is a valid projection under the Frobenius norm. \citet{sedghi2018singular} demonstrates a method to perform an approximation of the optimal projection to the orthogonal kernel space. Unfortunately, this method needs many expensive iterations to enforce the Lipschitz constraint tightly, making this approach computationally intractable in training large networks with \emph{provable} Lipschitz constraints. 

\paragraph{Comparison to BCOP} OSSN and SVCM's run-time depend on the input's spatial dimensions, which prohibits scalability (See Appendix \ref{appendix:algo-comp} for a time complexity analysis). RKO does not guarantee an exact Lipschitz constant, which may cause a loss in expressive power. Additionally, none of these methods guarantee gradient norm preservation. BCOP avoids all of the issues above.

\subsection{Provable Adversarial Robustness} 
Certifying the adversarial robustness of a network subject to norm ball perturbation is difficult. Exact certification methods using mixed-integer linear programming or SMT solvers scale poorly with the complexity of the network~\citep{katz2017reluplex, cheng2017maximum}. \citet{cohen2019certified} and \citet{salman2019provably} use an estimated smoothed classifier to achieve very high provable robustness with high confidence. In this work, we are primarily interested in providing deterministic provable robustness guarantees.

Recent work has been focusing on guiding the training of the network to be verified or certified (providing a lower-bound on provable robustness) easier~\citep{wong2018scaling, xiao2018training, croce2018provable, croce2019provable, hein2017formal}. For example, \citet{xiao2018training} encourage weight sparsity and perform network pruning to speed up the exact verification process for ReLU networks. \citet{wong2018scaling} optimize the network directly towards a robustness lower-bound using a dual optimization formulation. 

Alternatively, rather than modifying the optimization objective to incentivize robust classification, one can train networks to have a small global Lipschitz constant, which allows an easy way to certify robustness via the output margin. \citet{cohen2019universal} deploy spectral norm regularization on weight matrices of a fully connected network to constrain the Lipschitz constant and certify the robustness of the network at the test time. \citet{tsuzuku2018lipschitz} estimate an upper-bound of the network's Lipschitz constant and train the network to maximize the output margin using a modified softmax objective function according to the estimated Lipschitz constant. In contrast to these approaches, \citet{anil2018sorting} train fully connected networks that have a known Lipschitz constant by enforcing gradient norm preservation. Our work extends this idea to convolutional networks.

\section{Experiments}
The primary point of interest for the BCOP method (Section \ref{sec:lip-conv-bcop}) is its expressiveness compared against other common approaches of paramterizing Lipschitz constrained convolutions (Section \ref{sec:lip-conv-exist}). To study this, we perform an ablation study on two tasks using these architectures: The first task is provably robust image classification tasks on two datasets (MNIST~\citep{lecun1998mnist} and CIFAR10~\citep{cifar10})\footnote{We only claim in the deterministic case as recent approaches have much higher probabilistic provable robustness~\citep{cohen2019certified, salman2019provably}.}. We find our method outperformed other Lipschitz constrained convolutions under the same architectures as well as the state-of-the-art in this task (Section \ref{sec:exp-adv}). The second task is 1-Wasserstein distance estimation of GANs where our method also outperformed other competing Lipschitz-convolutions under the same architecutre (Section \ref{sec:exp-wde}).

\subsection{Network Architectures and Training Details}
A benefit of training GNP networks is that we enjoy the property of dynamical isometry, which inherently affords greater training stability, thereby reducing the need for common techniques that would otherwise be difficult to incorporate into a GNP network. For example, if a 1-Lipschitz residual connection maintains GNP, the residual block must be an identity function with a constant bias (see an informal justification in Appendix \ref{sec:res-con}). Also, batch normalization involves scaling the layer's output, which is not necessarily 1-Lipschitz, let alone GNP. For these reasons, residual connections and batch normalization are not included in the model architecture. We also use cyclic padding to substitute zero-padding since zero-padded orthogonal convolutions must be size 1 (see an informal proof in Appendix \ref{sec:zero-pad}). Finally, we use ``invertible downsampling''~\citep{jacobsen2018revnet} in replacement of striding and pooling to achieve spatial downsampling while maintaining the GNP property. The details for these architectural decisions are in Appendix~\ref{sec:net-details}.

Because of these architectural constraints, we base our networks on architectures that do not involve residual connections. For provable robustness experiments, we use the ``Small'' and ``Large'' convolutional networks from \citet{wong2018scaling}. For Wasserstein distance estimation, we use the fully convolutional critic from \citet{radford2015unsupervised} (See Appendix \ref{appendix:networdarch}, \ref{appendix:training} for details). Unless specified otherwise, each experiment is repeated 5 times with mean and standard deviation reported.

\subsection{Provable Adversarial Robustness}
\label{sec:exp-adv}
\subsubsection{Robustness Evaluation}
For adversarial robustness evaluation, we use the $L_2$-norm-constrained threat model~\citep{carlini2019evaluating}, where the adversary is constrained to $L_2$-bounded perturbation with the $L_2$ norm constrained to be below $\epsilon$. We refer to \emph{clean accuracy} as the percentage of un-perturbed examples that are correctly classified and \emph{robust accuracy} as the percentage of examples that are guaranteed to be correctly classified under the threat model. We use the margin of the model prediction to determine a lower bound of the robust accuracy (as described in Section \ref{sec:background-adv}). We also evaluate the empirical robustness of our model around under two gradient-based attacks and two decision-based attacks: \textit{(i)} PGD attack with CW loss~\citep{madry2017towards, carlini2017towards}, \textit{(ii)} FGSM~\citep{szegedy2013intriguing}, \textit{(iii)} Boundary attack (BA)~\citep{brendel2017decision}, \textit{(iv)} Point-wise attack (PA)~\citep{schott2018robust}. Specifically, the gradient-based methods (\textit{(i)} and \textit{(ii)}) are done on the whole test dataset; the decision-based attacks (\textit{(iii)} and \textit{(iv)}) are done only on the first 100 test data points since they are expensive to run.\footnote{We use foolbox~\citep{rauber2017foolbox} for the two decision-based methods.}

\subsubsection{Comparison of Different Methods for Enforcing Spectral Norm of Convolution}
\label{sec:methodscomp}
We compare the performance of OSSN, RKO, SVCM, and BCOP on margin training for adversarial robustness on MNIST and CIFAR10. To make the comparison fair, we ensure all the methods have a tight Lipschitz constraint of 1. For OSSN, we use 10 power iterations and keep a running vector for each convolution layer to estimate the spectral norm and perform the projection during every forward pass. For SVCM, we perform the singular value clipping projection with 50 iterations after every 100 gradient updates to ensure the Lipschitz bound is tight. For RKO, instead of using a regularization term to enforce the orthogonality (as done in \citet{cisse2017parseval}), we use Bj{\"o}rck~\citep{bjorck1971iterative} to orthogonalize the reshaped matrix before scaling down the kernel. We train two different convolutional architectures with the four aforementioned methods of enforcing Lipschitz convolution layers on image classification tasks. To achieve large output margins, we use first-order, multi-class, hinge loss with a margin of $2.12$ on MNIST and $0.7071$ on CIFAR10.

\begin{figure}
    \centering
    \includegraphics[width=\linewidth]{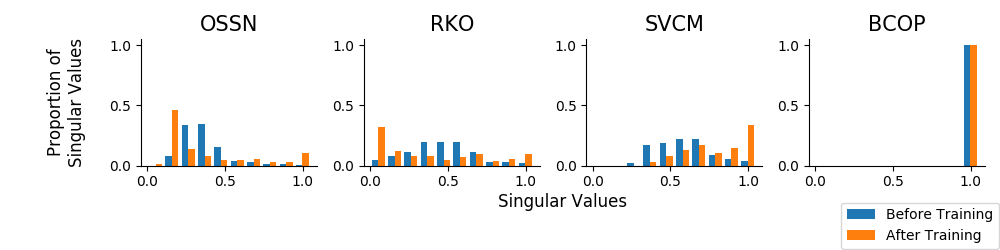}
    \caption{Singular value distribution at initialization (blue) and at the end of training (orange) for the second layer of the ``Large'' baseline using different methods to enforce Lipschitz convolution.}
    \label{fig:sv-viz}
\end{figure}

Our approach (BCOP) outperforms all competing methods across all architectures on both MNIST and CIFAR10 (See Table \ref{tb:mnist} and Appendix \ref{appendix:add-ablation-exps}, Table \ref{tb:additional}). To understand the performance gap, we visualize the singular value distribution of a convolution layer before and after training in Figure ~\ref{fig:sv-viz}. We observe that OSSN and RKO push many singular values to 0, suggesting that the convolution layer is not fully utilizing the expressive power it is capable of. This observation is consistent with our hypothesis that these methods bias the convolution operators towards sub-optimal regions caused by the loose Lipschitz bound (for RKO) and improper projection (for OSSN). In contrast, SVCM's singular values started mostly near 0.5 and some of them were pushed up towards 1, which is consistent with the procedure being an optimal projection. BCOP has all of its singular values at 1 throughout training by design due to its gradient norm preservation and orthogonality. Thus, we empirically verify the downsides of other methods and show that our proposed method enables maximally expressive Lipschitz constrained convolutional layers with guaranteed gradient-norm-preservation. 

\begin{table}[t!]
\centering
\begin{tabularx}{\textwidth}{c|cc|cccc}%
\textbf{Dataset}&\textbf{}&\textbf{}&\textbf{OSSN}&\textbf{RKO}&\textbf{SVCM}&\textbf{BCOP}\\%
\cmidrule{1-7}%
\multirow{4}{*}{\shortstack[c]{\textbf{MNIST} \\ ($\epsilon=1.58$)}}&\multirow{2}{*}{\textbf{Small}}&Clean&$96.86\pm0.13$&$97.28\pm0.08$&$97.24\pm0.09$&$\mathbf{97.54}\pm0.06$\\%
&&Robust&$42.95\pm1.09$&$43.58\pm0.44$&$28.94\pm1.58$&$\mathbf{45.84}\pm0.90$\\%
\cmidrule{2-7}%
&\multirow{2}{*}{\textbf{Large}}&Clean&$98.31\pm0.03$&$98.44\pm0.05$&$97.93\pm0.05$&$\mathbf{98.69}\pm0.01$\\%
&&Robust&$53.77\pm1.02$&$55.18\pm0.46$&$38.00\pm1.82$&$\mathbf{56.37}\pm0.33$\\%
\cmidrule{1-7}%
\multirow{4}{*}{\shortstack[c]{\textbf{CIFAR10} \\ ($\epsilon=36/255$)}}&\multirow{2}{*}{\textbf{Small}}&Clean&$62.18\pm0.66$&$61.77\pm0.63$&$62.39\pm0.46$&$\mathbf{64.53}\pm0.30$\\%
&&Robust&$48.03\pm0.54$&$47.46\pm0.53$&$47.59\pm0.56$&$\mathbf{50.01}\pm0.21$\\%
\cmidrule{2-7}%
&\multirow{2}{*}{\textbf{Large}}&Clean&$67.51\pm0.47$&$70.01\pm0.26$&$69.65\pm0.38$&$\mathbf{72.16}\pm0.23$\\%
&&Robust&$53.64\pm0.49$&$55.76\pm0.16$&$53.61\pm0.51$&$\mathbf{58.26}\pm0.17$\\%
\cmidrule{1-7}%
\end{tabularx}

\caption{Clean and robust accuracy on MNIST and CIFAR10 using different Lipschitz convolutions. The provable robust accuracy is evaluated at $\epsilon=1.58$ for MNIST and at $\epsilon=36/255$ for CIFAR10. }
\label{tb:mnist}
\end{table}

\subsubsection{State-of-the-art Comparison}
To further demonstrate the expressive power of orthogonal convolution, we compare our networks with models that achieve state-of-the-art deterministic provable adversarial robustness performance (Table \ref{tb:sota} and Appendix~\ref{appendix:add-exps}, Table~\ref{tb:sota-mmr} and Table~\ref{tb:sota-qw}). We also evaluate the empirical robustness of our model against common attacks on CIFAR10. Comparing against \citet{wong2018scaling}, our approach reaches similar performance for ``Small'' architecture and better performance for ``Large'' architecture (Table \ref{tb:cifar10-attack}).

\begin{table*}[t!]
\begin{center}
\begin{tabularx}{0.9\textwidth}{c|c|cccc}%
\textbf{Dataset}&&\textbf{BCOP-Large}&\textbf{FC-3}&\textbf{KW-Large}&\textbf{KW-Resnet}\\%
\cmidrule{1-6}%
\multirow{2}{*}{\shortstack[c]{\textbf{MNIST} \\ ($\epsilon=1.58$)}}&Clean&$98.69\pm0.01$&$\mathbf{98.71}\pm0.02$& $88.12$ &{-}\\%
&Robust&$\mathbf{56.37}\pm0.33$&$54.46\pm0.30$& $44.53$ &{-}\\%
\cmidrule{1-6}%
\multirow{2}{*}{\shortstack[c]{\textbf{CIFAR10} \\ ($\epsilon=36/255$)}}&Clean&$\mathbf{72.16}\pm0.23$&$62.60\pm0.39$& $59.76$ & $61.20$\\%
&Robust&$\mathbf{58.26}\pm0.17$&$49.97\pm0.35$&$50.60$ & $51.96$\\%
\cmidrule{1-6}%
\end{tabularx}
\end{center}
\caption{Comparison of our convolutional networks and the fully connected baseline in \citet{anil2018sorting} (FC-3) against provably robust models in previous works. The numbers for KW-Large and KW-Resnet are directly obtained from Table 4 of in the Appendix of their paper~\citep{wong2018scaling}.}
\label{tb:sota}
\end{table*}

\begin{table*}[t]
\centering
\parbox{.45\linewidth}{
\begin{tabularx}{\textwidth}{lcc}%
\textbf{}&\textbf{KW}&\textbf{BCOP}\\%
\cmidrule{1-3}%
\textbf{Small}&&\\%
\cmidrule{1-3}%
Clean&$54.39$&$\mathbf{64.53}\pm0.30$\\%
PGD&$49.94$&$\mathbf{51.26}\pm0.17$\\%
FGSM&$49.98$&$\mathbf{51.57}\pm0.18$\\%
\cmidrule{1-3}%
\textbf{Large}&&\\%
\cmidrule{1-3}%
Clean&$60.14$&$\mathbf{72.16}\pm0.23$\\%
PGD&$55.53$&$\mathbf{64.12}\pm0.13$\\%
FGSM&$55.55$&$\mathbf{64.24}\pm0.12$\\%
\end{tabularx}%
}
\parbox{.45\linewidth}{
\begin{tabularx}{\textwidth}{lcc}%
\textbf{}&\textbf{KW}&\textbf{BCOP}\\%
\cmidrule{1-3}%
\textbf{Small}&&\\%
\cmidrule{1-3}%
Clean (*)&$63.00$&$\mathbf{74.20}\pm2.23$\\%
BA (*)&$60.00$&$\mathbf{61.20}\pm2.99$\\%
PA (*)&$63.00$&$\mathbf{74.00}\pm2.28$\\%
\cmidrule{1-3}%
\textbf{Large}&&\\%
\cmidrule{1-3}%
Clean (*)&$68.00$&$\mathbf{78.40}\pm1.96$\\%
BA (*)&$64.00$&$\mathbf{71.00}\pm2.28$\\%
PA (*)&$68.00$&$\mathbf{78.40}\pm1.96$\\%
\end{tabularx}%
}

\caption{Comparison of our networks with \citet{wong2018scaling} on CIFAR10 dataset. Left: results of the evaluation on the entire CIFAR10 test dataset. Right (*): results of the evaluation on the first 100 test samples. The KW models~\citep{wong2018scaling} are directly taken from their official repository.}
\label{tb:cifar10-attack}
\end{table*}

\subsection{Wasserstein Distance Estimation}
\label{sec:exp-wde}
In this section, we consider the problem of estimating the Wasserstein distance between two high dimensional distributions using neural networks. \citet{anil2018sorting} showed that in the fully connected setting, ensuring gradient norm preservation is critical for obtaining tighter lower bounds on the Wasserstein distance. We observe the same phenomenon in the convolutional setting. 

We trained our networks to estimate the Wasserstein distance between the data and generator distributions of a GAN\footnote{Note that any GAN variant could have been chosen here.} \citep{goodfellow2014generative} trained on RGB images from the STL-10 dataset, resized to 64x64. After training the GAN, we froze the weights of the generator and trained Lipschitz constrained convolutional networks to estimate the Wasserstein distance. We adapted the fully convolutional discriminator model used by \citet{radford2015unsupervised} by removing all batch normalization layers and replacing all vanilla convolutional layers with Lipschitz candidates (BCOP, RKO, and OSSN)\footnote{We omit SVCM for comparison in this case due to its computational intractability}. We trained each model with ReLU or MaxMin activations \citep{anil2018sorting}. The results are shown in Table \ref{tb:wasserstein}.

\begin{table}[t!]
\begin{center}
\begin{tabularx}{0.85\textwidth}{ccccc}%
\textbf{}&\textbf{}&\textbf{OSSN}&\textbf{RKO}&\textbf{BCOP}\\%
\cmidrule{1-5}%
\multirow{2}{*}{\textbf{{Wasserstein Distance}}}&\textbf{MaxMin}&$7.39\pm0.31$&$8.95\pm0.12$&$\mathbf{9.91}\pm0.11$\\%
&\textbf{ReLU}&$7.06\pm0.72$&$7.82\pm0.21$&$\mathbf{8.28}\pm0.19$\\%
\end{tabularx}
\end{center}

\caption{Comparison of different Lipschitz constrained architectures on the Wasserstein distance estimation task between the data and generator distributions of an STL-10 GAN. Each estimate is a strict lower bound (estimated using 6,400 pairs of randomly sampled real and generated image examples), hence larger values indicate better performance.}
\label{tb:wasserstein}
\end{table}

Baking in gradient norm preservation in the architecture leads to significantly tighter lower bounds on the Wasserstein distance. The only architecture that has gradient norm preserving layers throughout (BCOP with MaxMin) leads to the best estimate. Although OSSN has the freedom to learn orthogonal kernels, this does not happen in practice and leads to poorer expressive power.

\section{Conclusion and Future Work}
We introduced convolutional GNP networks with an efficient construction method of orthogonal convolutions (BCOP) that overcomes the common issues of Lipschitz constrained networks such as loose Lipschitz bounds and gradient norm attenuation. In addition, we showed the space of orthogonal convolutions has many connected components and demonstrated how BCOP parameterization alleviates the optimization challenges caused by the disconnectedness. Our GNP networks outperformed the state-of-the-art for deterministic provable adversarial robustness on $L_2$ metrics with both CIFAR10 and MNIST, and obtained tighter Wasserstein distance estimates between high dimensional distributions than competing approaches. Despite its effectiveness, our parameterization is limited to only expressing a subspace of orthogonal convolutions. A complete parameterization of the orthogonal convolution space may enable training even more powerful GNP convolutional networks. We presented potential directions to achieve this and left the problem for future work.

\subsubsection*{Acknowledgements}
We would like to thank Lechao Xiao, Arthur Rabinovich, Matt Koster, and Siqi Zhou for their valuable insights and feedback.

\bibliographystyle{plainnat}
\bibliography{references}
\newpage
\appendix
\section{Optimizing under spectral normalization}
\label{app:optimizing_sn}

Here we provide theoretical analysis of the optimization properties of spectral normalization. We focus on the setting in which the weight matrices are projected to the feasible set via spectral normalization after each gradient update (i.e. projected gradient descent).

Firstly, we note that spectral normalization is a valid projection under the operator 2-norm \citep{gouk2018regularisation} but not the Frobenius norm, where the projection would clip all singular values larger than 1 \citep{lefkimmiatis2013hessian}. Despite this, all existing implementations of spectral normalization as a projection perform steepest descent optimization in Euclidean space which is not guaranteed to converge \citep{bertsekas1997nonlinear}. We illustrate this with a simple example.

\paragraph{Spectral norm projection counter-example} Consider a constraint optimization problem: \begin{equation}
    A^* = \argmax_{A: ||A||_2\leq1} \{  \Tr(AD) \},
\end{equation} where $A$ and $D$ are diagonal, with $\diag(D) = [2, 1]$. Clearly, the objective is maximized by $A^*=I$. However, the Euclidean steepest ascent direction is given by the gradient, which is $D$ in this case. A single gradient update (with learning rate $\alpha$) and projection step acting on the diagonal of A looks like this (assuming that $x + 2\alpha > y + \alpha$ throughout the course of learning, which is indeed the case given the initialization):

\begin{equation}
    \begin{bmatrix}
         x \\
         y 
    \end{bmatrix} \leftarrow
    \begin{bmatrix}
         \min\{ x + 2\alpha, 1 \} \\
         (y + \alpha) / \max\{x + 2\alpha, 1\} 
    \end{bmatrix}
\end{equation}

This update eventually converges to $\diag(A) = [1, 0.5]$, not the identity.

How do we fix this? To make sure that projected gradient descent will converge to the correct stationary point we must choose our descent direction to induce the most change under the correct norm: the operator 2-norm. Doing so leads to the following update,
\begin{lemmasec}[Steepest descent by matrix 2-norm]
The first order approximation of the steepest descent direction under the matrix operator 2-norm is given by the gradient with all non-zero singular values rescaled to be equal. That is, given a loss function $L: \mathbb{R}^{n\times m} \rightarrow \mathbb{R}$, and corresponding gradient at $W$, $G = \nabla L(W) = U\Lambda V^T$, one steepest descent direction is $\bar{D} = -U \mathcal{P}(\Lambda) V^T$, where the projection operator $\mathcal{P}$ sets all non-zero elements of $\Lambda$ to 1.
\label{lemma:2norm_steepest_descent}
\end{lemmasec}

\begin{proof} \textbf{(Lemma \ref{lemma:2norm_steepest_descent})}
We seek the steepest descent direction,

\begin{equation*}
    \bar{D} = \argmin_D \{ L(W + D) \: : \: ||D||_2 = 1 \}
\end{equation*}

Consider the first-order Taylor expansion of the loss,

\begin{equation*}
L(W + \bar{D}) \approx L(W) + \Tr(G\bar{D}^T),
\end{equation*}

where the trace is computing the vectorized dot product between the gradient and the descent direction. Thus, the first order approximation of the steepest descent direction seeks to minimize $\Tr(G\bar{D}^T)$ subject to the 2-norm constraint on $D$. Without loss of generality, we will write $D = UU' S V'^TV^T$, then we wish to minimize $\Tr(V' S U'^T \Lambda) = \Tr(K \Lambda) = \sum_i \lambda_i K_{ii}$, where we write $K=V'SU'^T$, and $\lambda_i$ denotes the diagonal elements of $\Lambda$. We have reduced this to a simple constrained optimization problem where we wish to make $K_{ii}$ as negative as possible. This can be achieved when $K_{ii}=-1$ for every $\lambda_i \neq 0$. Thus, we have $\bar{D} = -U \mathcal{P}(\Lambda) V^T$.
\end{proof}

In our experiments, we did not use the projection step with the correct steepest descent direction but instead opted to rescale the matrices by their spectral norm during the forward pass and backpropagate through this step.

\section{Examples of Lipschitz Functions}
\label{appendix:lip-examples}
\paragraph{Affine Transformations} All affine transformations are Lipschitz functions, and their Lipschitz constant under $L_2$ is given by their spectral norm, which is the largest singular value of that linear transformation. A noteworthy special subset of linear transformations is the subset of orthogonal linear transformations. A linear transformation is considered orthogonal if it has maximal rank and all non-zero singular values equal to 1. They have the special property that an orthogonal transformation $O: \mathbb{R}^n \mapsto \mathbb{R}^m$  will preserve norm if $n \leq m$, that is, $||Ox||_2 = ||x||_2, \forall x$. In the backward pass (backpropagation), the gradient signal becomes $O^Tg$, where $g$ is is the incoming gradient. Since the transpose of an orthogonal linear transformation is also orthogonal, we see that $O$ is gradient-norm-preserving for $n \geq m$.

\paragraph{GroupSort} GroupSort is a 1-Lipschitz activation function that is proposed in \citet{anil2018sorting}. GroupSort partitions the activation vector into groups of same size and sorts the values within each group in-place. \citet{anil2018sorting} showed that GroupSort can recover ReLU and the absolute value function. Most importantly, it addresses the capacity limitation induced by ReLU activation in Lipschitz constrained networks~\citep{huster2018limitations}.

\section{Algorithm Complexity of Different Approaches to Enforce Lipschitz Convolution}
\label{appendix:algo-comp}

For simplicity, we assume all the matrices to be square matrices with a size of $c \times c$ and convolution kernel to be $k \times k \times c \times c$ where $k$ is the kernel size and $c$ is the channel size. We also assume all the input has a spatial dimension of $s \times s$. In practice, the kernel size of the convolution is usually small (e.g., $k=3$), so we consider it as a constant factor. Thus, we are mainly interested in considering the time complexity with respect to $s$ and $c$. In addition, we assume matrix multiplication of two $c \times c$ matrices takes $O(c^3)$.

\paragraph{Orthogonalization using Bj{\"o}rck and Power iteration} First order Bj{\"o}rck orthogonalization on a $c \times c$ matrix takes $O(c^3)$ per iteration. We use power iterations to rescale the matrix before the orthogonalization procedure to ensure convergence, which take $O(c^2)$ per iteration. Overall, the orthogonalization takes $O(c^3)$. In practice, we use 20 iterations of first order Bj{\"o}rck and 10 power iterations.

\paragraph{OSSN} This method computes an approximated spectral norm for the convolution operator and scales down by that value. The approximated spectral norm is computed by power iteration for convolution \citep{gouk2018regularisation}, which involves  convolving the convolution kernel on a tensor with the full input shape and convolving the transposed convolution kernel\footnote{Convolving a transposed kernel is equivalent to applying a transposed linear transformation} on the full output shape per iteration. Overall, the time complexity is $O(c^3s^2)$.

\paragraph{RKO} This method simply orthogonalizes an $c \times k^2c$ matrix, so it takes O($c^3$).

\paragraph{SVCM} Singular value clipping and masking takes $O(c^3s^2)$ per iteration (as analyzed in \citet{sedghi2018singular}).

\paragraph{BCOP} BCOP requires one orthogonalization of a $c \times c$ matrix and $2k-2$ orthogonalizations of $c \times \left\lceil\frac{c}{2}\right\rceil$ matrices for the symmetric projectors. Overall, the runtime for BCOP is $O(c^3)$. 

\begin{table}[hbt!]
\begin{center}
\begin{tabularx}{0.42\linewidth}{cccc}
    \textbf{Standard} & \textbf{BCOP} & \textbf{RKO} & \textbf{OSSN}\\ \midrule
    0.041 & 0.138 & 0.120 & 0.113
\end{tabularx}
\end{center}
\caption{Time (in seconds) that each of the method takes for one training iteration with the ``Large'' architecture on one NVIDIA P100 GPU on CIFAR10 dataset. A batch size of 128 is used.}
\label{tb:time}
\end{table}

\section{Architectural Detail Considerations for GNP Convolutional Networks}
\label{sec:net-details}
In our construction of Lipschitz convolutional networks, we restrict ourselves further to GNP convolutional networks. This has the benefit of preventing the gradient norm attenuation when each layer of a Lipschitz network is constrained, as well as giving training stability through dynamical isometry. To build a GNP network, we make every component of the network GNP. GNP convolutions have already been established by using BCOP to make orthogonal convolutions; however, there are still many more elements of a convolutional network to make GNP. An important realization is that due to dynamical isometry in GNP networks, it is no longer necessary to use the typical methods for adding stability in training, so these elements may be removed from the network. The following will discuss all the architectural decisions made for constructing networks with the GNP property while also leveraging the properties that GNP affords.

\subsection{Residual Connections}
\label{sec:res-con}
Residual connections make it difficult and unnatural to maintain a small Lipschitz constant for the network while being GNP. As well, a key feature of residual connections yields a stabler Jacobian for better training dynamics; however, the dynamical isometry property of the networks means that additional stability is not necessary. Therefore, we remove residual connections from our Lipschitz convolutional network designs. 

A residual connection layer can be expressed as $g(x) = f(x)+x$, where $f$ is generally parameterized by some shallow or deep neural network. We can then bound the Lipschitz constant $\Lip(g)$ in terms of $\Lip(f)$,
\begin{align*}
    ||g(x_1)-g(x_2)|| = ||(f(x_1)+x_1)-(f(x_2)+x_2)|| \leq ||f(x_1)-f(x_2)|| + ||x_1-x_2||
\end{align*}
So we have $\Lip(g) \leq 1+\Lip(f)$, which may be a loose bound in general, but a tighter bound is not easy to determine. To guarantee that $g$ is 1-Lipschitz, we could only do so by having $\Lip(f)=0$, which means $f$ is a constant function, which obviously is not sufficiently expressive. Therefore, getting a class of 1-Lipschitz functions with residual connections is not straightforward. A possible workaround to this could be to constrain the Lipschitz constant of $f$ to 1, then halve the value after the residual connection, i.e. $g(x) = \frac{f(x)+x}{2}$. This indeed will bound $\Lip(g)$ by 1, but then a problem arises with gradient norm preservation. The Jacobian of $g$ would be 
\begin{align*}
    \nabla_x g = \frac{\nabla_x f + I}{2}
\end{align*}
So for $g$ to be GNP almost everywhere, we would require $\nabla_x g = \frac{\nabla_x f + I}{2}$ to be orthogonal almost everywhere; however, there is no natural or well-known way to optimize over a class of non-linear functions $f$ such that $\frac{\nabla_x f + I}{2}$ is orthogonal almost everywhere. These reasons show why residual connections are hard to reconcile with Lipschitz-constrained and GNP networks.

\subsection{Zero-Padded Orthogonal Convolutions}
\label{sec:zero-pad}
Consider an orthogonal 1-D convolution kernel of size $2k+1$, represented by
\begin{align*}
    [A_{-k}\ \cdots\ A_0\ \cdots\ A_k]
\end{align*}
Then the corresponding Toeplitz matrix of the zero-padded convolution operation is
\begin{align*}
    M = \begin{pmatrix}
        A_0 & A_1 & \hdots & A_k & 0 & \hdots & 0 & \hdots & 0 \\
        \vdots & \vdots & \ddots & \vdots & \vdots & \ddots & \vdots & \ddots & \vdots \\
        A_{-k} & A_{-k+1} & \hdots & A_0 & A_1 & \hdots & 0 & \hdots & 0 \\
        \vdots & \vdots & \ddots & \vdots & \vdots & \ddots & \vdots & \ddots & \vdots \\
        0 & 0 & \hdots & 0 & 0 & \hdots & A_{-k} & \hdots & A_0
    \end{pmatrix}
\end{align*}

Since the kernel is orthogonal, $MM^T = I$. This means that if $R_i$ is the $i^{th}$ block row of $M$, then $R_iR_j^T = \delta_{ij}I$. In particular, we can take the first block row, then $(k+1)^{th}$ block row (i.e. the one with $A_{-k}$ as the first element), and the last row. The first row yields the condition
\begin{align*}
    \sum_{i=0}^k A_iA_i^T = I
\end{align*}
The $(k+1)^{th}$ row yields
\begin{align*}
    \sum_{i=-k}^k A_iA_i^T = I
\end{align*}
And the last row yields
\begin{align*}
    \sum_{i=-k}^0 A_iA_i^T = I
\end{align*}
Combining these conditions yields that $A_0A_0^T = I$. This then implies that all other matrices must be 0. Therefore, all 1-D orthogonal kernels with zero-padding are only size 1 kernels, and so cyclic padding is used instead.

\subsection{Invertible Downsampling}
\label{appendix:invdownsampling}
\begin{figure}
    \centering
    \includegraphics[width=.4\linewidth]{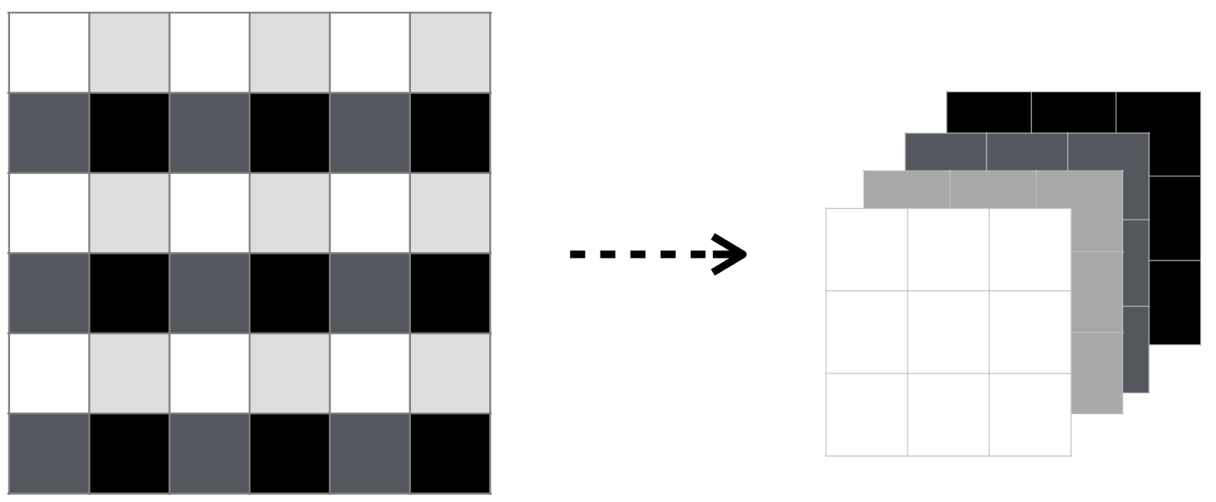}
    \caption{Invertible Downsampling~\citep{jacobsen2018revnet}}
    \label{fig:i_ds}
\end{figure}

The theory developed for the orthogonal convolution assumed stride 1. As such, we make sure all the convolutions are done with only stride of 1. However, since striding is an important feature in convolutional networks, we emulate it through an \textbf{invertible downsampling} layer~\citep{jacobsen2018revnet} followed by a stride-1 convolution. Invertible downsampling rearranges pixels in a single channel into multiple channels so that a stride-1 convolution over the rearranged image is equivalent to a strided convolution over the original image. This layer is illustrated in Figure \ref{fig:i_ds} with input channel size 1 and stride 2. 

\subsection{Other Components}
\textbf{Batch Normalization} Batch normalization is generally used to improve stability in training; however, it is neither 1-Lipschitz nor gradient norm preserving. Therefore, it is removed from the network.

\textbf{Linear Layers} We directly use Bj{\"o}rck orthogonalization (See Section \ref{appendix:ortho-procedure}) procedure to enforce orthogonal linear layers as done in \citet{anil2018sorting}.

\textbf{Activation Functions} The activation function we use is GroupSort as \citet{anil2018sorting} found that GroupSort enhances the network capacity of Lipschitz networks compared against ReLU. In particular, we use GroupSort with a group size of 2, which is referred to as MaxMin~\citep{anil2018sorting} (or OPLU in \citet{chernodub2016norm}). We use MaxMin activation because we found it to work the best in practice.

\section{Network Architectures}
\label{appendix:networdarch}
We describe the details for the network architectures we used in this paper and compare the number of parameters (See Table \ref{tb:num-params}).

\paragraph{Small}
The ``Small'' convolutional network contains two convolutional layers with kernel size of 4, stride 2, and channel sizes of 16 and 32 respectively, followed by two linear layers with 100 hidden units.

\paragraph{Large}
The ``Large'' convolutional network contains four convolutional layers with kernel size of 3/4/3/4 and stride 1/2/1/2 with channel sizes of 32/32/64/64 respectively, followed by three linear layers with 512 hidden units.

\paragraph{FC-3}
The ``FC-3'' networks refer to a 3-layer fully connected network with the number of hidden units of $1024$.

\begin{table}[hbt!]
\begin{center}
\begin{tabularx}{\textwidth}{ccccc}
    & \textbf{Small} & \textbf{Large} & \textbf{FC-3} & \textbf{DCGAN Critic} \\ \midrule
    \textbf{MNIST} & 166,406 & 1,974,762 & 2,913,290 & \\
    \textbf{CIFAR10} & 214,918 & 2,466,858 & 5,256,202 & \\
    \textbf{Wasserstein Distance Estimation} & & & & 2,764,737
\end{tabularx}
\end{center}
\caption{Number of parameters for each architecture on different tasks.}
\label{tb:num-params}
\end{table}

\paragraph{DCGAN Critic} All of the Wassertein distance estimation experiments uses a variant of the fully convolutional critic architecture described by \citet{radford2015unsupervised}. This architecture consists of 5 convolutional layers with kernel sizes of 4/4/4/4/4, strides of 2/2/2/2/1 and channel sizes of 64/128/256/512/1. We removed all the Batch Normalization layers and used either ReLU activation or MaxMin activation. 

It is important to note that, in general, it is difficult to make the whole network gradient-norm-preserved because a linear transformation from a low dimensional vector to a high dimensional vector is guaranteed to lose gradient norm under some inputs. Since the aforementioned architectures mostly consist of layers that are decreasing in size, enforcing orthogonality is sufficient to enforce gradient norm preservation throughout most part of the networks (usually only the first layer is increasing in dimension). 

\section{Training Details}
\label{appendix:training}
\paragraph{Provable Robustness for MNIST and CIFAR10}
We used Adam optimizer and performed a search over $0.01$, $0.001$, $0.0001$ for the learning rate. We found $0.001$ to work the best for all experiments. We used exponential learning decay at the rate of $0.1$ per $60$ epochs. We trained the networks for 200 epochs with a batch size of 128. All of our experiments were run on NVIDIA P100 GPUs. No preprocessing was done on MNIST dataset (pixel values are between 0 and 1). For CIFAR10, standard data augmentation is applied with random cropping (with a maximum padding of 4 pixels) and random horizontal flipping. The pixel values are between 0 and 1 with no scaling applied.

\paragraph{Wasserstein Distance Estimation}
The STL-10 GANs we used were trained using the gradient penalized Wasserstein GAN framework  \citep{arjovsky2017wasserstein, gulrajani2017improved}. The generator and discriminator network architectures were adapted from the ones used by \citet{chen2016infogan}. The implementation as well as the choice of hyperparameters is based on \cite{gan_impl}. A learning rate of 0.0002 was used for both the generator and discriminator. The gradient penalty applied on the discriminator was 50. The model was trained for 128 epoch, with a batch size of 64 using the Adam optimizer \citep{kingma2014adam} with $\beta_{1} = 0.5$ and $\beta_{2}=0.999$. The training was performed on NVIDIA P100 GPUs. 

The DCGAN architecture \cite{radford2015unsupervised} was used to independently compute the Wasserstein distance between the data and generator distributions of the aforementioned GAN. We used RMSprop optimizer and performed a search over $0.1, 0.01, 0.001, 0.0001$ for the learning rate. We found that learning rate of $0.001$ works the best for models with ReLU activation and learning rate of $0.0001$ works the best MaxMin activation. The numbers reported in the table were corresponding to the best learning rates. In practice, we also observed that the training with OSSN can be unstable under high learning rate in our Wasserstein distance estimation experiments. We also used the same exponential learning decay and a batch size of 64. All the networks were trained for 25,600 iterations on NVIDIA P100 GPUs.

\section{Orthogonalization Procedure}
\label{appendix:ortho-procedure}
Several ways have been proposed in the literature to orthogonalize an unconstrained matrix in a differentiable manner~\citep{pmlr-v97-lezcano-casado19a, bjorck1971iterative}. In this work, we adopt Bj{\"o}rck orthogonalization algorithm from \citet{bjorck1971iterative}.

The original Bj{\"o}rck paper proposes the following iterative procedure to find the closest matrix under the metric of the Frobenius norm of the difference matrix:
\begin{align*}
    \text{Bj{\"o}rck}(A) = A\left(I + \frac{1}{2}Q + \frac{3}{8}Q^2 + \dots + (-1)^p {-\frac{1}{2}\choose p}Q^p\right)
\end{align*}
where $p \geq 1$ and $Q = I-A^TA$. This function can be iterated arbitrarily to get tighter estimates of orthogonal matrices. In all our experiments, we use $p=1$ and iteratively apply this function 20 times.

\section{BCOP Implementation}
\label{appendix:bcop-procedure}
BCOP consists of a series of block convolutions with symmetric projectors in each of the convolution component. In practice, one could use any unconstrained matrix $R  \in \mathbb{R}^{n \times \lfloor \frac{n}{2} \rfloor}$ to parameterize a symmetric projector $P \in \mathbb{R}^{n \times n}$ with $\rank(P) = \lfloor \frac{n}{2} \rfloor$ as follows:
\begin{equation}
\begin{aligned}
    \tilde{R} &= \mathrm{Bj\text{\"o}rck}(R) \\
    P &= \tilde{R}\tilde{R}^T
\end{aligned}
\end{equation}
where $\mathrm{Bj\text{\"o}rck}$ standards for the Bj\"orck orthongalization algorithm that computes the closest orthogonal matrix (closeness in terms of the Frobenius norm) given an arbitrary input matrix~\citep{bjorck1971iterative} (See Appendix \ref{appendix:ortho-procedure} for details on orthogonalization procedure). 

For convergence guarantees of Bj{\"o}rck, we also rescale the unconstrained matrix to be spectral norm bounded by 1 using power iteration. This rescaling procedure does not change the output orthogonal matrix at convergence because Bj{\"o}rck is scale-invariant~\citep{bjorck1971iterative}, i.e., $\mathrm{Bj\text{\"o}rck}(\alpha R) = \mathrm{Bj\text{\"o}rck}(R)$.

\section{Additional Ablation Experiments}
\label{appendix:add-ablation-exps}
In this section, we report the performance of some other alternative Lipschitz constrained convolutions and compare them against BCOP's performance (Table \ref{tb:additional}). 

\paragraph{BCOP-Fixed} Same as BCOP method (as introduced in Section \ref{sec:lip-conv-bcop}) but the weights of the convolutions were frozen during training. Only the weights in the fully-connected layer are being optimized. This method was tested as a sanity check to ensure that BCOP isn't offloading all the training to the fully connected layer while the BCOP convolutional layers did little of the work, as was a phenomenon observed in \citet{abadi2016deep,cox2011beyond}. 

\paragraph{RK-L2NE} Another alternative reshaped kernel (RK) method that bounds the spectral norm of a matrix as done in \citet{qian2018l2}. In particular,
\begin{align*}
    ||A||_2 \leq \max(||AA^T||_\infty, ||A^TA||_\infty) 
\end{align*}
We first compute the upper-bound of the spectral norm of the reshaped kernel as above and then scale the matrix down by the factor. Similar to RKO, we scale the convolution kernel (reshaped from the matrix that has a spectral norm of at most 1) down by a factor of $K$ with $K$ being the kernel size of the convolution to ensure the spectral norm of the convolution is not greater than 1. All of the computations above are done during the forward pass.

\begin{table}[t!]
\centering
\begin{tabularx}{0.9\textwidth}{c|cc|ccc}%
\textbf{Dataset}&\textbf{}&\textbf{}&\textbf{BCOP-Fixed}&\textbf{RK-L2NE}&\textbf{BCOP}\\%
\cmidrule{1-6}%
\multirow{4}{*}{\shortstack[c]{\textbf{MNIST} \\ ($\epsilon=1.58$)}}&\multirow{2}{*}{\textbf{Small}}&Clean&$93.57\pm0.17$&$95.85\pm0.12$&$\mathbf{97.54}\pm0.06$\\%
&&Robust&$7.51\pm1.18$&$39.77\pm0.73$&$\mathbf{45.84}\pm0.90$\\%
\cmidrule{2-6}%
&\multirow{2}{*}{\textbf{Large}}&Clean&$69.12\pm5.61$&$96.76\pm0.11$&$\mathbf{98.69}\pm0.01$\\%
&&Robust&$0.00\pm0.00$&$37.79\pm1.21$&$\mathbf{56.37}\pm0.33$\\%
\cmidrule{1-6}%
\multirow{4}{*}{\shortstack[c]{\textbf{CIFAR10} \\ ($\epsilon=36/255$)}}&\multirow{2}{*}{\textbf{Small}}&Clean&$50.61\pm0.65$&$58.82\pm0.67$&$\mathbf{64.53}\pm0.30$\\%
&&Robust&$36.44\pm0.70$&$44.65\pm0.61$&$\mathbf{50.01}\pm0.21$\\%
\cmidrule{2-6}%
&\multirow{2}{*}{\textbf{Large}}&Clean&$47.70\pm0.92$&$56.75\pm0.68$&$\mathbf{72.16}\pm0.23$\\%
&&Robust&$27.47\pm1.37$&$43.40\pm0.46$&$\mathbf{58.26}\pm0.17$\\%
\cmidrule{1-6}%
\end{tabularx}

\caption{Clean and robust accuracy on MNIST and CIFAR10 using different Lipschitz convolutions. The provable robust accuracy is evaluated at $\epsilon=1.58$ for MNIST and at $\epsilon=36/255$ for CIFAR10. Each experiment is repeated 5 times.}
\label{tb:additional}
\end{table}

\begin{table*}[!ht]
\begin{center}
\begin{tabularx}{0.8\textwidth}{c|c|ccc}%
\textbf{Dataset}&&\textbf{BCOP-Large}&\textbf{FC-3}&\textbf{MMR-Universal}\\%
\cmidrule{1-5}%
\multirow{2}{*}{\shortstack[c]{\textbf{MNIST} \\ ($\epsilon=0.3$)}}&Clean&$98.69\pm0.01$&$\mathbf{98.71}\pm0.02$&{96.96}\\%
&Robust&$97.04\pm0.03$&$\mathbf{97.06}\pm0.02$&{89.60}\\%
\cmidrule{1-5}%
\multirow{2}{*}{\shortstack[c]{\textbf{CIFAR10} \\ ($\epsilon=0.1$)}}&Clean&$\mathbf{72.16}\pm0.23$&$62.60\pm0.39$&{53.04}\\%
&Robust&$\mathbf{62.53}\pm0.21$&$53.67\pm0.29$&{46.40}\\%
\cmidrule{1-5}%
\end{tabularx}
\caption{Comparison of our convolutional networks against the provable robust model from \citet{croce2019provable}. The numbers for MMR-Universal are directly obtained from Table 1 in their paper~\citep{croce2019provable}.}
\label{tb:sota-mmr}
\end{center}
\end{table*}

\begin{table*}[!ht]
\begin{center}
\begin{tabularx}{0.8\textwidth}{c|c|cccc}%
\textbf{Dataset}&&\textbf{BCOP-Large}&\textbf{FC-3}&\textbf{QW-3}&\textbf{QW-4}\\%
\cmidrule{1-6}%
\multirow{4}{*}{\shortstack[c]{\textbf{MNIST} \\ ($\epsilon=1.58$)}}&Clean&$98.69\pm0.01$&$\mathbf{98.71}\pm0.02$&$98.65$&$98.23$\\%
&Robust&$\mathbf{56.37}\pm0.33$&$54.46\pm0.30$&$42.13$&$27.59$\\%
&PGD&$86.38\pm0.16$&$81.96\pm0.16$&$\mathbf{86.86}$&$86.25$\\%
&FGSM&$\mathbf{86.49}\pm0.16$&$83.64\pm0.10$&$85.83$&$84.17$\\%
\cmidrule{1-6}%
\multirow{4}{*}{\shortstack[c]{\textbf{CIFAR10} \\ ($\epsilon=36/255$)}}&Clean&$72.16\pm0.23$&$62.60\pm0.39$&$\mathbf{79.15}$&$77.15$\\%
&Robust&$\mathbf{58.26}\pm0.17$&$49.97\pm0.35$&$44.46$&$31.41$\\%
&PGD&$64.12\pm0.13$&$50.05\pm0.36$&$\mathbf{72.07}$&$71.89$\\%
&FGSM&$64.24\pm0.12$&$50.21\pm0.34$&$\mathbf{72.11}$&$71.92$\\%
\cmidrule{1-6}%
\end{tabularx}
\caption{Comparison of our convolutional networks against the provable robust models (QW-3 for Model-3 and QW-4 for Model-4) from \citet{qian2018l2}. The model weights are directly downloaded from their official website.}
\label{tb:sota-qw}
\end{center}
\end{table*}

\section{Comparison to Other Baselines for Provable Adversarial Robustness on MNIST and CIFAR10}
\label{appendix:add-exps}
We also compare the performance of the GNP networks with another baseline (MMR-Universal) from \citet{croce2019provable}. The provable robustness results are summarized in Table \ref{tb:sota-mmr}. Both our ``Small'' and ``Large'' model outperform the ``Small'' model from \citet{croce2019provable} in terms of clean accuracy and robust accuracy. The fully connected Lipschitz constrained network baseline from \citet{anil2018sorting} achieves slightly better performance than convolutional networks we use for MNIST, but does much worse for CIFAR10. 

In addition, we also compare against the models from \citet{qian2018l2}. To encourage robustness against adversary, they proposed to use (norm) non-expansive operation only, which equivalently enforces the Lipschitz constant of the network to be at most 1. Instead of using a single network to predict the logits for all the classes, \citet{qian2018l2} uses a separate 1-Lipschitz network to predict the logit for each of the 10 classes in both MNIST and CIFAR10 datasets. It can be shown that we can at best certify $\bx$ is robustly classified if
\begin{align}
\label{eq:qw-criterion}
    \mathcal{M}_f(\bx) > 2\epsilon.
\end{align}
We report the provably robust accuracy (using the certification criterion presented in Equation \ref{eq:qw-criterion}) and the robust accuracy against two gradient-based attacks in Table \ref{tb:sota-qw}. It is important to note that the QW models are designed to be robust under empirical attacks instead of obtaining certification of robustness. Also, the perturbation sizes used in their original paper are much larger than the ones that we are focusing on.

\section{Topology of of 1-D Orthogonal Convolution Kernel}
\label{appendix:disconnectedness}

In this section, we introduce the following theorem:
\onedbcop*
To prove this theorem, we first discuss important properties of symmetric projectors which is of central importance of the proof in Appendix \ref{appendix:background-sym}. Followed by the discussion, we focus on finding the connected components of a subset of orthogonal convolution kernels -- special orthogonal convolution kernels (SOCK) -- in Appendix \ref{appendix:sock}. Finally, we show how the connected components of SOCK can be trivially extended to the connected components of orthogonal convolution kernel.

\subsection{Background: Symmetric Projector}
\label{appendix:background-sym}
Before we discuss the topology of orthogonal convolution kernels, we first review some basic properties of symmetric projectors. 

\begin{definition} $\mathbb{P}(n)$ is the space of all $n\times n$ symmetric projectors. Formally, $$\mathbb{P}(n) = \{P | P^2 = P^T = P, P \in \mathbb{R}^{n\times n}\}$$ 
We also denote the space of rank-$k$ symmetric projectors as $\mathbb{P}(n, k)$:
\begin{align*}
    \mathbb{P}(n, k) = \{P | \rank(P) = k, P \in \mathbb{P}(n)\}
\end{align*}\label{def:sym-proj}\end{definition}

\begin{remarkdef} Followed from the definition of symmetric projectors, we can make a few observations:
\begin{enumerate}
    \item Each symmetric projector can be identified with an orthogonal projection onto a linear subspace. 
    \item The range operator of matrix is a bijection map between $\mathbb{P}(n)$ and all linear subspaces of $\mathbb{R}^n$. In particular, the map bijectively sends $\mathbb{P}(n,k)$ to $\Gr(k, \mathbb{R}^n)$ (or Grassmannian manifold), which is defined as the set of all $k$-dimensional subspaces of $\mathbb{R}^n$~\citep{milnor1974characteristic}. 
\end{enumerate}
\end{remarkdef}

\begin{theorem}[Symmetric Projectors and Grassmannian Manifold]
$\mathbb{P}(n, k)$ and $\Gr(k, \mathbb{R}^n)$ are homeomorphic.
\end{theorem}

\begin{remark} The homeomorphism allows us to inherit properties from Grassmannian manifold:
    \begin{enumerate}
        \item $\mathbb{P}(n,k)$ is compact and path connected
        \item $\mathbb{P}(n,k)$ is disjoint from $\mathbb{P}(n,k')$ for $k\neq k'$
        \item By compactness and disjointness above, $\mathbb{P}(n,k) \cup \mathbb{P}(n,k')$ for $k \neq k'$ is path disconnected.
        \item The dimensionality of $\mathbb{P}(n,k)$ is $k(n-k)$, which is maximized when $k=\lceil \frac{n}{2} \rceil,\lfloor \frac{n}{2} \rfloor$.
    \end{enumerate}
\label{remark:sym}
\end{remark}

\subsection{Connected Components of 1-D Special Orthogonal Convolution Kernels (SOCK)}
\label{appendix:sock}
We will be using the results discussed in the previous section to identify the connected components of orthogonal convolution kernel (OCK) submanifold. 
In the rest of this section, we put our focus on the case of special orthogonal convolution kernel (SOCK) where the convolution operator belongs to the special orthogonal group. Because of the symmetry between the orthogonal transformation with $\det = 1$ and $\det = -1$, we can trivially determine all the connected components of OCK from the connected components of SOCK as there is no intersection between the components of OCK with $\det(H) = -1$ and $\det(H) = 1$. For the rest of this section, we put our focus on a representation of SOCK:

\begin{definition} A \textbf{1-D special orthogonal convolution kernel (SOCK) submanifold}, denoted by $\mathcal{C}(r_1, r_2, \cdots, r_{K-1})$, is a submanifold of $\mathbb{R}^{n \times nK}$ that can be represented by

\begin{equation}
    A = H \square \begin{bmatrix}P_1 & I-P_1\end{bmatrix} \square \cdots \square \begin{bmatrix} P_{K-1} & I-P_{K-1}\end{bmatrix}
    \label{eq:construction}
\end{equation}
where 
\begin{itemize}
    \item ``$\square$'' is the block convolution operator that convolves one block matrix with another:
        \begin{equation*}
            \begin{bmatrix}X_1 & X_2 & \cdots & X_p \end{bmatrix} \square \begin{bmatrix} Y_1 & Y_2 & \cdots & Y_q \end{bmatrix} = \begin{bmatrix}Z_1 & Z_2 & \cdots & Z_{p+q-1} \end{bmatrix}
        \end{equation*} with $Z_i = \sum_j X_j Y_{i-j} $, where the out-of-range elements being all zero (e.g., $X_{<1} = 0, X_{>p} = 0, Y_{<1} = 0, Y_{>q} = 0$).
    \item $P_i \in \mathbb{P}(n, r_i), \forall i$. We refer $r = (r_1, r_2, \cdots, r_{K-1})$ as the rank tuple of the SOCK submanifold $\mathcal{C}(r_1, r_2, \cdots, r_{K-1})$, or $\mathcal{C}(r)$ in short.
    \item $H \in SO(n)$. 
\end{itemize}

We shorthand the representation described above as $A = \mathcal{A}(P_1, \cdots, P_{K-1}, H)$. Formally, 
\begin{equation*}
    \mathcal{C}(r) = \mathcal{C}(r_1, r_2, \cdots, r_{K-1}) = \{A | A = \mathcal{A}(P_1, \cdots, P_{K-1}, H), P_i \in \mathbb{P}(n, r_i), H \in SO(n)\}.
\end{equation*}
We can also define
\begin{equation*}
    \mathcal{C} = \{A | A = \mathcal{A}(P_1, \cdots, P_{K-1}, H), P_i \in \mathbb{P}(n), H \in SO(n)\}.
\end{equation*}
\end{definition}

\begin{theorem}[Completeness (Theorem 2 of \citet{kautsky1994matrix})]
    $\mathcal{C}$ is the space of 1-D special orthogonal convolution kernels.
\end{theorem}

\begin{definition} The \textbf{canonical rank tuple} of special orthogonal convolution is defined as
\begin{align*}
    r^{(k)} = \left(r^{(k)}_1, r^{(k)}_2, \cdots, r^{(k)}_{K-1}\right)
\end{align*}
with the following conditions:
\begin{enumerate}
    \item $\sum_i r^{(k)}_i = k$
    \item $\left|r^{(k)}_i - r^{(k)}_{i'}\right| \leq 1, \forall i, i'$
    \item $r^{(k)}_i \leq r^{(k)}_{i+1}, \forall i$
\end{enumerate}

Intuitively, these conditions enforce the ranks to be most balanced while having their sum equal to $k$. The last condition makes $r^{(k)}$ unique for each $k$. Since each rank tuple defines a SOCK submanifold, we can define the \textbf{canonical SOCK submanifold} as follows:
\begin{align*}
    \mathcal{C}_k = \mathcal{C}\left(r^{(k)}\right) = \mathcal{C}\left(r^{(k)}_1, r^{(k)}_2, \cdots, r^{(k)}_{K-1}\right).
\end{align*}
\label{def:cr}
\end{definition}

\begin{theorem}[Space of 1-D Convolution Kernel]
1-D special orthogonal convolution space is compact and it consists of $(K-1)n+1$ distinct canonical SOCK submanifolds as its connected components: $\mathcal{C}_0, \mathcal{C}_1, \cdots, \mathcal{C}_{(K-1)n}$.
\label{thm:1d-conv-space-general}  
\end{theorem}

The main idea of proving this theorem is to show that \textbf{any SOCK with the sum of its rank tuple equal to $k$ is a subset of $\mathcal{C}_k$}. Our proof of the theorem is divided into the following three steps:
\begin{enumerate}
    \item \textbf{Equivalent SOCK Construction}: We identify the changes that we can make to the rank of the symmetric projectors in the representation (Equation \ref{eq:construction}) without changing the kernel that they represent, and find pairs of SOCK submanifolds in which one fully contains another (Appendix \ref{appendix:eq-sock-cons}).
    \item \textbf{Dominance of Canonical SOCK Submanifold}: We prove that the canonical SOCK submanifolds fully contain all other SOCK submanifolds using the relationship between SOCK submanifolds identified above. A consequence of this result is that the union of the canonical SOCK submanifolds $\mathcal{C}_k$ is complete (Appendix \ref{appendix:canonical-sock-dominates}).
    \item \textbf{Connected Components of $\mathcal{C}$ are Canonical SOCK submanifolds}: Given the result in Step 2, we complete the proof of Theorem \ref{thm:1d-conv-space-general} by showing that the canonical SOCK submanifolds $\mathcal{C}_k$ are compact and disjoint, and hence the number of connected components of $\mathcal{C}$ is $(K-1)n+1$, which is the number of distinct canonical SOCK submanifolds. (Appendix \ref{appendix:conn-c-are-sock-submanifolds}).
\end{enumerate}

\subsubsection{Equivalent SOCK Construction}
\label{appendix:eq-sock-cons}
Now, we introduce one important property of symmetric projectors that guide the construction of equivalent representations with changes in the symmetric projectors.
\begin{lemma}[Symmetric Projector Pair Equivalence under Product and Sum]
For all $P'_1 \in \mathbb{P}(n,k_1-1)$ and $P'_2 \in \mathbb{P}(n,k_2+1)$ with $1\leq k_1 \leq k_2+1$, there always exists $P_1 \in \mathbb{P}(n,k_1)$ and $P_2 \in \mathbb{P}(n,k_2)$ such that $P'_1 + P'_2 = P_1 + P_2, P'_1P'_2 = P_1P_2, P'_2P'_1 = P_2P_1$. 
\label{lemma:half-rank-dominates}
\end{lemma}

\begin{proof}\textbf{(Lemma \ref{lemma:half-rank-dominates})} 
Let $Q'_1$ be the range of $P'_1$ and $Q'_2$ be the range of $P'_2$. We observe that $\dim(Q'_1+Q^{'\perp}_2) \leq \dim(Q'_1) + \dim(Q^{'\perp}_2) = k_1 - 1 + n - k_2 - 1 \leq n - 1$. Therefore, there always exists a unit vector $\bx$ in $Q'_2 \cap Q^{'\perp}_1$. Then, we can find a orthonormal basis decomposition of $Q'_2$ that contains $\bx$: $Q'_2 = \spn(\{\bx_1, \bx_2, \cdots, \bx_{k_2}, \bx\})$. Then, we can construct the linear subspaces $Q_1$ and $Q_2$ as follows:
\begin{align}
    Q_1 &= Q'_1 + \spn(\{\bx\}) \\
    Q_2 &= \spn(\{\bx_1, \bx_2, \cdots, \bx_{k_2}\}) \label{eq:q2}
\end{align}
Now, we can define $P_1$ and $P_2$ to be the symmetric projectors whose range are $Q_1$ and $Q_2$ respectively. By the construction of $\bx$, it is clear that $Q_1$ has one more dimension than $Q'_1$ and $Q_2$ has one less dimension than $Q'_2$, which makes $\rank(P_1) = k_1$ and $\rank(P_2) = k_2$. We are then only left to prove that $P_1 + P_2 = P'_1 + P'_2$, $P'_1P'_2 = P_1P_2$, and $P'_2P'_1 = P_2P_1$. 

We first observe that the orthogonal projection $P_1$ can be decomposed into the sum of two orthogonal projections onto two orthogonal subspaces: $P_1 = P'_1 + \bx\bx^T$ where the first projection is onto $Q'_1$ and the second projection is onto $\spn(\{\bx\})$ with $Q'_1 \perp \spn(\{\bx\})$. Similarily, $Q_2$ and $\spn(\{\bx\})$ are orthogonal subspaces and $Q'_2 = Q_2 + \bx$ from Equation \ref{eq:q2}. Decomposing $P'_2$ leads to $P_2 = P'_2 - \bx\bx^T$. 

From here it is clear that
\begin{align*}
    P_1 + P_2 &= P'_1 + \bx\bx^T + P'_2 - \bx\bx^T = P'_1 + P'_2 
\end{align*}

Since $\bx \in Q^{'\perp}_1$ and $\bx \in Q'_2$, we have $P'_1\bx = \mathbf{0}, P'_2\bx = \bx$. This allows us to complete the the proof:
\begin{align*}
    P_1P_2 &= (P'_1 + \bx\bx^T)(P'_2 - \bx\bx^T)
    = P'_1P'_2 + \bx(P'_2\bx)^T - \bx\bx^T
    = P'_1P'_2, \\
    P_2P_1 &= (P'_2 - \bx\bx^T)(P'_1 + \bx\bx^T)
    = P'_2P'_1 + \bx\bx^T - \bx(P'_1\bx)^T - \bx\bx^T
    = P'_2P'_1
\end{align*}
\end{proof}

\begin{restatable}[Equivalent SOCK Construction]{lemma}{eqk}
$\mathcal{A}(P_1, \cdots, P_i, P_{i+1}, \cdots, P_{K-1}, H) = \mathcal{A}(P_1, \cdots, P'_i, P'_{i+1}, \cdots, P_{K-1}, H')$ iff $P_{i} + P_{i+1} = P'_{i} + P'_{i+1}$, $P_iP_{i+1} = P'_iP'_{i+1}$, and $H = H'$. 
\label{lemma:eq_kernel_construction_k}
\end{restatable}

The proof of Lemma \ref{lemma:eq_kernel_construction_k} is in Appendix \ref{appendix:prop-sym-proj}.

Now, we can find pairs of SOCK submanifold where one fully contains another.
\begin{lemma}[Balanced Rank Dominates]$\mathcal{C}(r_1, r_2, \cdots, r_{i} - 1, r_{i+1} + 1, \cdots r_{K-1}) \subseteq \mathcal{C}(r_1, r_2, \cdots r_{K-1})$ when $r_{i} \leq r_{i+1}+1$ and 
$\mathcal{C}(r_1, r_2, \cdots, r_{i} + 1, r_{i+1} - 1, \cdots r_{K-1}) \subseteq \mathcal{C}(r_1, r_2, \cdots r_{K-1})$ when $r_{i+1} \leq r_{i}+1$. 
\label{lemma:balance-rank-dominates}
\end{lemma}

\begin{proof} 
In the case where $r_i \leq r_{i+1}+1$, suppose $P_i \in \mathbb{P}(n,r_i-1)$ and $P_{i+1} \in \mathbb{P}(n,r_{i+1}+1)$. By Lemma \ref{lemma:half-rank-dominates}, there exists $P'_i \in \mathbb{P}(n,r_i)$ and $P'_{i+1} \in \mathbb{P}(n,r_{i+1})$ such that $P_{i} + P_{i+1} = P'_{i} + P'_{i+1}$, $P_iP_{i+1} = P'_iP'_{i+1}$. By Lemma \ref{lemma:eq_kernel_construction_k}, $\mathcal{A}(P_1,\cdots,P_{K-1},H) = \mathcal{A}(P_1,\cdots,P'_i,P'_{i+1}, \cdots P_{K-1},H)$. Since this holds regardless of the choice of $P$'s, we can conclude $\mathcal{C}(r_1, r_2, \cdots, r_{i} - 1, r_{i+1} + 1, \cdots r_{K-1}) \subseteq \mathcal{C}(r_1, r_2, \cdots r_{K-1})$. 

In the case where $r_{i+1} \leq r_i+1$, the same proof holds by symmetry.
\end{proof}

\begin{lemma}[Rank Balancing]
$\mathcal{C}(r_1, r_2, \cdots, r_{i} - \delta, \cdots, r_{j} + \delta, \cdots,  r_{K-1}) \subseteq \mathcal{C}(r_1, r_2, \cdots, r_{i}, \cdots, r_{j}, \cdots,  r_{K-1})$ if $r_{i} = r_{p} = r_{j}$ for $i < p < j$ and $\delta \in \{-1, 1\}$.
\label{lemma:sequence-rank-balance}
\end{lemma}

\begin{proof} 
The Lemma can be proven by induction on $i$ facilitated by Lemma \ref{lemma:balance-rank-dominates}.
\end{proof}

\subsubsection{Dominance of Canonical SOCK Submanifold}
\label{appendix:canonical-sock-dominates}
Using all the subspace relations that we identify above (Lemma \ref{lemma:balance-rank-dominates} and \ref{lemma:sequence-rank-balance}) repeatedly from any SOCK submanifold, we can find another SOCK submanifold with a more balanced rank tuple that fully contains the former submanifold. To formalize the notion of ``balanced rank tuple'', we introduce \textbf{imbalance score function}:

\begin{definition} The imbalance score for an special orthogonal convolution subspace $\mathcal{C}(r)$ is
\begin{equation}
    f(r) = \sum_{i} (r_i)^2
\end{equation}
For any rank tuple that has the minimum imbalance score under the constraint that the sum of them is $k$, we call it a \textbf{balanced rank tuple} with sum $k$. It is clear that the condition of having the minimum imbalance score is also equivalent to having the following condition:
\begin{align*}
    \left|r_i - r_j\right| \leq 1, \forall i, j
\end{align*}
\label{def:imbalance-score}
\end{definition}

\begin{lemma}[SOCKs with Balanced Rank Tuple are Equiavlent]
Let $r$ be a balanced rank tuple. Then, $\mathcal{C}(r) = \mathcal{C}_k$ with $k = \sum_i r_i$.
\label{lemma:balanced-rank-equivalent}
\end{lemma}
\begin{proof}
Any balanced rank tuple has a rank difference of at most 1. Then, we can use Lemma \ref{lemma:balance-rank-dominates} to swap any two adjacent ranks such that
\begin{align*}
    \mathcal{C}(r_1, \cdots, c_{i+1}, c_{i}, \cdots, r_{K-1}) \subseteq \mathcal{C}(r_1, \cdots, c_{i}, c_{i+1}, \cdots, r_{K-1}).
\end{align*}
Now, if we apply the swap again, we will get
\begin{align*}
    \mathcal{C}(r_1, \cdots, c_{i+1}, c_{i}, \cdots, r_{K-1}) \supseteq \mathcal{C}(r_1, \cdots, c_{i}, c_{i+1}, \cdots, r_{K-1}),
\end{align*}
which means 
\begin{align*}
    \mathcal{C}(r_1, \cdots, c_{i+1}, c_{i}, \cdots, r_{K-1}) = \mathcal{C}(r_1, \cdots, c_{i}, c_{i+1}, \cdots, r_{K-1}).
\end{align*}
From here, it is obvious that we can propagate the equivalance relationship from any balanced rank tuple to a canonical rank tuple by performing bubble sort while the sum of the ranks remains the same. Therefore, any SOCK submanifold with its sum of ranks equals to $k$ would be equivalent to $\mathcal{C}_k$. 
\end{proof}

Since a SOCK with balanced rank tuple is equiavlent to its corresponding canonical SOCK, we are only left to show that the imbalance score function can always be decreased until the rank is most balanced. 

\begin{lemma}[Balancing Subspace Dominates]
Given a rank tuple $r$, if there exists $i, j$ such that $\left| r_i - r_j\right| \geq 2$, then there exists $\mathcal{C}(r') \supseteq \mathcal{C}(r)$ with $\sum_i r'_i = \sum_i r_i$ such that $f(r) > f(r')$.
\label{lemma:balance-sub-dom}
\end{lemma}

\begin{proof}\textbf{(Lemma \ref{lemma:balance-sub-dom})}
Let $i<j$ be the closest pair of points such that $|r_i-r_j|\geq 2$. Without loss of generality, we assume $r_i < r_j$ since the same proof below would hold for the case with $r_i > r_j$ by symmetry. Now, consider the following rank tuple
\begin{align*}
    r' = (r_1, \cdots, r_i + 1, \cdots, r_j - 1, \cdots, r_{K-1}).
\end{align*}
It is clear that
\begin{align*}
    f(r') - f(r) = 2(r_i - r_j) < 0.
\end{align*}
Now, we are left to show that $\mathcal{C}(r') \supseteq \mathcal{C}(r)$.

\paragraph{Case 1} Assume the two ranks are adjacent, or $i+1 = j$. By Lemma \ref{lemma:balance-rank-dominates}, $\mathcal{C}(r) \subseteq \mathcal{C}(r')$.
\paragraph{Case 2} Assume the two ranks are not adjacent, or $i+1<j$. We must have $r_p=r_i+1=r_j-1$ for $i<p<j$. Otherwise, the $i$ and $j$ would no longer be the closest pair such that $|r_i-r_j|\geq 2$. We can then apply Lemma \ref{lemma:sequence-rank-balance} to get $\mathcal{C}(r) \subseteq \mathcal{C}(r')$.
\end{proof}

\begin{lemma}[Canonical Submanifold Dominates]
$\mathcal{C}(r') \subseteq \mathcal{C}_k$ if $\sum_{p} r'_p = k$.
\label{lemma:balance-rank-dominates-2}
\end{lemma}

\begin{proof}\textbf{(Lemma 
\ref{lemma:balance-rank-dominates-2})} We prove this Lemma by dividing it up into the following two cases:
\paragraph{Case 1}
Assume there is no $i, j$ such that $|r'_i - r'_j| \geq 2$, the rank tuple $r'$ is balanced by Definition \ref{def:imbalance-score}. Then, by Lemma \ref{lemma:balanced-rank-equivalent}, $\mathcal{C}(r') = \mathcal{C}_k$.

\paragraph{Case 2}
Assume there exists $i, j$ such that $|r'_i - r'_j| \geq 2$. Then by Lemma \ref{lemma:balance-sub-dom}, we can always find a special orthogonal convolution subspace that contains the current subspace $\mathcal{C}(r') $ without changing the sum of the ranks and with the strictly decreased imbalance score. Since the imbalance score takes on natural numbers, iteratively applying the Lemma to decrease the imbalance score must eventually terminate. When it does, it will yield the subspace $\mathcal{C}(r^*)$ which contains the $\mathcal{C}(r')$ and has $|r^*_i - r^*_j| < 2$ for all $i,j$. We are left to show that $\mathcal{C}(r^*) \subseteq \mathcal{C}_k$, which is shown in the first case.
\end{proof}

\begin{corollary}[The union of all canonical subspaces is complete ]
$\mathcal{C} = \mathcal{C}_0 \cup \mathcal{C}_1 \cup \cdots \cup \mathcal{C}_{(K-1)n}$.
\label{corollary:balance-rank-dominates-3}
\end{corollary}

\begin{proof}\textbf{(Corollary \ref{corollary:balance-rank-dominates-3})}
$\mathcal{C}$ is the union of all SOCK submanifolds, where each of them belong to at least one of $\mathcal{C}_k$ by Lemma \ref{lemma:balance-rank-dominates-2}. From here, it is clear that the Corollary holds.
\end{proof}

\subsubsection{Connected Components of $\mathcal{C}$ are Canonical SOCK Submanifolds}
\label{appendix:conn-c-are-sock-submanifolds}

To fill in the final piece of proving Theorem \ref{thm:1d-conv-space-general}, we need to prove that $\mathcal{C}_k$ are disjoint and compact. We first prove the disjointness by expressing the sum of symmetric projectors used to construct the kernel as a linear combination of kernel elements $\{A_1, A_2, \cdots, A_K\}$ below. The proof of Lemma \ref{lemma:a-component} is in Appendix \ref{appendix:prop-sym-proj}.
\begin{restatable}[Kernel Element Decomposition]{lemma}{copa}
    If $A=\mathcal{A}(P_1, \cdots, P_{K-1},H)$, then $A_j = \sum_{i=0}^j a_{K,j,i} B_i$, where $a_{K,j,i} =  (-1)^{j-i} \binom{(K-1)-i}{j-i}$ for $i \leq j \leq K-1$ and 0 otherwise, and $B_k = \sum_{{\delta'_1, \delta'_2, \cdots, \delta'_{K-1}} \mid \sum_i \delta'_i=k} H\prod_{1\leq i\leq K-1}\left[(1-\delta'_i)P_i + \delta'_iI\right]$
    \label{lemma:a-component}
\end{restatable}

\begin{corollary}[Triangular Map between $A$ and $B$]
    $B_j = A_j - \sum_{k=0}^{j-1} (-1)^{j-k}\binom{(K-1)-k}{j-k} B_k$. 
    \label{corollary:sum-equal}
\end{corollary}

\begin{proof}\textbf{(Corollary \ref{corollary:sum-equal})}
The expression can be obtained by simply rearrange the expression of $A_j$ from Lemma \ref{lemma:a-component}.
\end{proof}

\begin{proof}\textbf{(Theorem \ref{thm:1d-conv-space-general})}
From Corollary \ref{corollary:sum-equal}, we can recursively expand out the terms on the right to express $B_j$ as a linear combination of $\{H^TA_1, H^TA_2, \cdots, H^TA_{K-1}\}$. Formally, $B_j = \sum_k (w_{jk} H^T A_k)$ for some $w_{ji}$. We can then define a continuous function $g$, $g(A) = \Tr \sum_k (w_i H^T A_i) = \Tr H^T B_{K-2} = \sum_i \Tr (H^THP_i) =  \sum_i \Tr P_i = \sum_i r_i = r$. Therefore $g(\mathcal{C}_r) = \{r\}$ and $g(\mathcal{C}_{r'}) = \{r'\}$, so $\mathcal{C}_r,\mathcal{C}_{r'}$ must be disjoint if $r\neq r'$.

Next, we prove compactness of each $\mathcal{C}(r)$. We first observe that symmetric projector submanifold, $\mathbb{P}(n, k)$, is compact and path connected for any $k$ (Remark \ref{remark:sym}). $\mathcal{C}(r)$ is the image of $\mathcal{A}$ under these $K-1$ sets of $P_i$'s and the set of special orthogonal matrices $H$'s. All of these sets are compact and path-connected and $\mathcal{A}$ is continuous; therefore $\mathcal{C}(r)$ is path connected and compact.

Finally, since $\mathcal{C}_r$ and $\mathcal{C}_{r'}$ are compact and disjoint, $\mathcal{C}_r \cup \mathcal{C}_{r'}$ is path-disconnected for $r\neq r'$. Combining this with the connectedness of each individual $\mathcal{C}_r$ as well as completeness of the $\{\mathcal{C}_r\}$ (Corollary \ref{corollary:balance-rank-dominates-3}), we can conclude that the $\mathcal{C}_r$'s are the connected components of $\mathcal{C}$, so there are $(K-1)n+1$ total disconnected components.
\end{proof}

\onedbcop*
\begin{proof}
From Theorem \ref{thm:1d-conv-space-general}, $C$ is the union of all SOCK submanifolds which contain $(K-1)n+1$ connected components. The other subset of orthogonal convolution kernels that we omitted when considering SOCK can be simply obtained by negating one row of the orthogonal matrix in the SOCK representation (Equation \ref{eq:construction}). Because there exists no continuous path from the components with determinants of $-1$ to the components with determinants of $1$. The number of connected components in orthogonal convolution kernels is therefore doubled, which is $2(K-1)n+2$. 
\end{proof}

\section{Additional Proofs}
\label{appendix:prop-sym-proj}
\eqk*
\begin{proof}\textbf{(Lemma 
\ref{lemma:eq_kernel_construction_k})}
Let
\begin{itemize}
    \item $A = \mathcal{A}(P_1, \cdots, P_i, P_{i+1}, \cdots, P_{K-1}, H)$
    \item $A' = \mathcal{A}(P_1, \cdots, P'_i, P'_{i+1}, \cdots, P_{K-1}, H')$
    \item $Q$ be a function of a set of binary variables ($\delta_j$) that control which factor ($P_j$ or $I - P_j$) appears in the product of matrices in the function output:
    \begin{equation*}
        Q(\delta_1, \delta_2, \cdots, \delta_{K-1}) = H\prod_{1\leq j\leq K-1}\left[(1-\delta_j)P_j + \delta_j(1-P_j)\right],
    \end{equation*}
    where $\delta_j \in \{0, 1\}$.
    \item $Q'$ be the function in similar form as $Q$, but the summation elements are constructed with respect to the new kernel $A'$.
\end{itemize}

Using the $Q$ function above, we can represent $A$ in an alternate form
\begin{equation*}
    A_j = \sum_{\delta_1, \delta_2, \cdots \delta_{K-1} \mid \sum_k \delta_k = j - 1} Q(\delta_1, \delta_2, \cdots, \delta_{K-1})
\end{equation*}
where $1 \leq j \leq K-1$. This form will make our proof below more convenient.

The proof is divided into the following two steps (forward direction and backward direction):
\paragraph{``$\Rightarrow$'' Direction}
\begin{equation*}
    A = A' \Rightarrow P_{i} + P_{i+1} = P'_{i} + P'_{i+1}, P_iP_{i+1} = P'_iP'_{i+1}, H = H'
\end{equation*}

We start by summing all the elements of $A$ and $A'$ to show $H = H'$:
\begin{equation*}
\begin{aligned}
    \sum_j A_j &= \sum_j \sum_{{\delta_1, \delta_2, \cdots, \delta_{K-1}} \mid \sum_k \delta_k = j - 1} Q(\delta_1, \delta_2, \cdots, \delta_{K-1}) \\ &= \sum_{{\delta_1, \delta_2, \cdots, \delta_{K-1}}} Q(\delta_1, \delta_2, \cdots, \delta_{K-1}) \\
    &= \sum_{{\delta_1, \delta_2, \cdots, \delta_{K-1}}} H\prod_{1\leq j\leq K-1}\left[(1-\delta_j)P_j + \delta_j(I-P_j)\right] \\
    &= H \prod_{1\leq j\leq K-1} [P_j + (I-P_j)] \\
    &= H
\end{aligned}
\end{equation*}
The same process for $A'$ will show $\sum_j A'_j = H'$. Thus, $H=H'$. We are still left to show $P_i+P_{i+1} = P'_i + P'_{i+1}$ and $P_iP_{i+1} = P'_iP'_{i+1}$. Now, we will consider $A$ and $A'$ in their alternate form,
\begin{align*}
    A &= H \square \begin{bmatrix} P_1 & I-P_1\end{bmatrix} \square \cdots \square \begin{bmatrix} P_{K-1} & I-P_{K-1}\end{bmatrix} \\
    A' &= H' \square \begin{bmatrix} P_1 & I-P_1\end{bmatrix} \square \cdots \begin{bmatrix} P'_{i} & I-P'_{i}\end{bmatrix} \square \begin{bmatrix} P'_{i+1} & I-P'_{i+1}\end{bmatrix} \square \cdots  \square \begin{bmatrix} P_{K-1} & I-P_{K-1}\end{bmatrix}
\end{align*}
First, we do a left block convolution by $H^T$ to obtain
\begin{align*}
    H^T \square A &= H^T \square H \square \begin{bmatrix} P_1 & I-P_1\end{bmatrix} \square \cdots  \square \begin{bmatrix} P_{K-1} & I-P_{K-1}\end{bmatrix} \\
    &= \begin{bmatrix} P_1 & I-P_1\end{bmatrix} \square \cdots  \square \begin{bmatrix} P_{K-1} & I-P_{K-1}\end{bmatrix} \\
    H^T \square A' &= H^T \square H' \square \begin{bmatrix} P_1 & I-P_1\end{bmatrix} \square \cdots \square\begin{bmatrix} P'_{i} & I-P'_{i}\end{bmatrix} \square \begin{bmatrix} P'_{i+1} & I-P'_{i+1}\end{bmatrix} \square \cdots  \square \begin{bmatrix} P_{K-1} & I-P_{K-1}\end{bmatrix} \\
    &= \begin{bmatrix} P_1 & I-P_1\end{bmatrix} \square \cdots \square\begin{bmatrix} P'_{i} & I-P'_{i}\end{bmatrix} \square \begin{bmatrix} P'_{i+1} & I-P'_{i+1}\end{bmatrix}\square \cdots  \square \begin{bmatrix} P_{K-1} & I-P_{K-1}\end{bmatrix}
\end{align*}
Now, since only the $i^{th}$ and $(i+1)^{th}$ convolutions differ between these sequences, we can iteratively convolve away every other convolution by left/right-convolving with the inverse of the left/right-most element. On the left, we would begin by convolving with $\begin{bmatrix} I-P_1 & P_1 \end{bmatrix}$ and on the right by $\begin{bmatrix} I-P_{K-1} & P_{K-1} \end{bmatrix}$. We can then continue performing the left/right convolution to repeatedly cancel out the left/right-most element until we are left with the two terms with index of $i$ and $i+1$ as follows:
\begin{align*}
    \begin{bmatrix} P_i & I - P_i \end{bmatrix} \square \begin{bmatrix} P_{i+1} & I - P_{i+1} \end{bmatrix} &= \begin{bmatrix} P'_i & I - P'_i \end{bmatrix} \square \begin{bmatrix} P'_{i+1} & I - P'_{i+1} \end{bmatrix}
\end{align*}

Expanding out the convolution and re-arranging the equation gives $P_i+P_{i+1} = P'_i + P'_{i+1}$ and $P_iP_{i+1} = P'_iP'_{i+1}$.

\paragraph{``$\Leftarrow$'' Direction}
\begin{equation*}
    P_{i} + P_{i+1} = P'_{i} + P'_{i+1}, P_iP_{i+1} = P'_iP'_{i+1}, H = H' \Rightarrow A=A'
\end{equation*}

To prove the backward direction, we can simply invert the proof in the forward direction:

$P_i+P_{i+1} = P'_i + P'_{i+1}$ and $P_iP_{i+1} = P'_iP'_{i+1}$ implies that
\begin{align*}
    \begin{bmatrix} P_i & I - P_i \end{bmatrix} \square \begin{bmatrix} P_{i+1} & I - P_{i+1} \end{bmatrix} &= \begin{bmatrix} P'_i & I - P'_i \end{bmatrix} \square \begin{bmatrix} P'_{i+1} & I - P'_{i+1} \end{bmatrix}
\end{align*}

Then, it is clear that
\begin{align*}
    &H \square \begin{bmatrix} P_1 & I-P_1\end{bmatrix} \square \cdots \square  \begin{bmatrix} P_{i} & I-P_{i}\end{bmatrix} \square \begin{bmatrix} P_{i+1} & I-P_{i+1}\end{bmatrix} \square \cdots  \square \begin{bmatrix} P_{K-1} & I-P_{K-1}\end{bmatrix} \\
    &= H' \square \begin{bmatrix} P_1 & I-P_1\end{bmatrix} \square \cdots \square\begin{bmatrix} P'_{i} & I-P'_{i}\end{bmatrix} \square \begin{bmatrix} P'_{i+1} & I-P'_{i+1}\end{bmatrix} \square \cdots  \square \begin{bmatrix} P_{K-1} & I-P_{K-1}\end{bmatrix},
\end{align*}
which yields $A = A'$ as what we needed.
\end{proof}

\copa*
\begin{proof} \textbf{(Lemma 
\ref{lemma:a-component})}

We will show this result by considering the following form of $A_j$,
    \begin{align*}
        A_j = \sum_{{\delta_1, \delta_2, \cdots, \delta_{K-1}} \mid \sum_i \delta_i = j} Q(\delta_1, \delta_2, \cdots, \delta_{K-1})
    \end{align*}
where $Q$ is given by
\begin{equation*}
    Q(\delta_1, \delta_2, \cdots, \delta_{K-1}) = H\prod_{1\leq i\leq K-1}\left[(1-\delta_i)P_i + \delta_i(I-P_i)\right]
\end{equation*}
and $\delta_i \in \{0, 1\}$.

Every summand of $A_j$ can be expanded into the form $$ Q(\delta_1, \delta_2, \cdots, \delta_{K-1}) = \sum_{{\delta'_1, \delta'_2, \cdots, \delta'_{K-1}} \mid \sum_i \delta'_i \leq j} a_{\delta'_1,\cdots,\delta'_{K-1}} H\prod_{1\leq i\leq K-1}\left[(1-\delta'_i)P_i + \delta'_iI\right]$$ for some coefficients $\{a_{\delta'_1,\cdots,\delta'_{K-1}}\}$, and we will provide a closed form for these coefficients.

When $\delta_i = 0$, the $i^{th}$ factor of $Q(\delta_1, \delta_2, \cdots, \delta_{K-1})$ is just $P_i$, and this term does not expand. This means that all summands in the expansion must contain $P_i$. Therefore, $a_{\delta'_1,\cdots,\delta'_{K-1}} = 0$ whenever $\delta'_i = 1$. When $\delta_i = 1$, the $i^{th}$ factor is instead $I-P_i$, which would generate two parts in the expansion, one with $I$, and the other with $-P_i$. Therefore, if $\delta_i = 1$, for a given $a_{\delta'_1,\cdots,\delta'_{K-1}}$, the $i^{th}$ factor provides a factor of -1 to the coefficient if $\delta'_i = 0$, and provides a factor of 1 otherwise. This means that the final coefficient will be $(-1)^n$, where $n$ is the number of positions $i$ where $\delta_i = 1$ but $\delta'_i = 0$. Thus the closed form is
\begin{equation*}
a_{\delta'_1,\cdots,\delta'_{K-1}} = 
    \begin{cases*}
        0, & \text{if } $\exists\ i$ \text{ such that } $\delta_i = 0, \delta'_i = 1$ \\
        (-1)^{\sum_i \delta_i - \delta'_i}, & \text{otherwise}
    \end{cases*}
\end{equation*}

Notice that $\sum_i \delta_i$ is constant for all summands of $A_j$, so $a_{\delta'_1,\cdots,\delta'_{K-1}}$ is constant over all summands in which it is non-zero. Therefore, we can find how many summands in which this coefficient is non-zero, and that will give us the total coefficient for $H\prod_{1\leq i\leq K-1}\left[(1-\delta'_i)P_i + \delta'_iI\right]$ in $A_j$.

To find this for a given $\{\delta'_i\}$, we simply need to find which $\{\delta_i\}$ satisfying $\sum_i \delta_i = j$ also satisfy ``$\forall i, \delta'_i = 1 \Rightarrow \delta_i = 1$.'' (which is just the negation of the property that leads to the coefficient being 0). We first start by letting $k = \sum_i \delta'_i$. All valid $\{\delta_i\}$ satisfy $\delta'_i = 1 \Rightarrow \delta_i = 1$, so the $i$ where $\delta'_i = 1$ forces $\delta_i = 1$. The final condition to satisfy is the sum. The forced positions sum to $k$, so it remains that the of the free positions, their total must sum to $j-k$. That is, out of the $(K-1)-k$ free positions, exactly $j-k$ of them must be 1. Clearly, this means there are $\binom{(K-1)-k}{j-k}$ $\{\delta_i\}$'s with a non-zero $a_{\delta'_1,\cdots,\delta'_{K-1}}$. Furthermore, $a_{\delta'_1,\cdots,\delta'_{K-1}} = (-1)^{j-k}$ for each. Therefore, we can combine all of these together to get
\begin{align*}
    A_j &= \sum_{{\delta'_1, \delta'_2, \cdots, \delta'_{K-1}} \mid \sum_i \delta'_i \leq j} (-1)^{j-\sum_i \delta'_i}\binom{n-\sum_i \delta'_i}{j-\sum_i \delta'_i} H\prod_{1\leq i\leq K-1}\left[(1-\delta'_i)P_i + \delta'_iI\right] \\
    &= \sum_{k=0}^j \sum_{{\delta'_1, \delta'_2, \cdots, \delta'_{K-1}} \mid \sum_i \delta'_i=k} (-1)^{j-k}\binom{(K-1)-k}{j-k} H\prod_{1\leq i\leq K-1}\left[(1-\delta'_i)P_i + \delta'_iI\right] \\
    &= \sum_{k=0}^j (-1)^{j-k}\binom{(K-1)-k}{j-k} \sum_{{\delta'_1, \delta'_2, \cdots, \delta'_{K-1}} \mid \sum_i \delta'_i=k} H\prod_{1\leq i\leq K-1}\left[(1-\delta'_i)P_i + \delta'_iI\right] \\
    &= \sum_{k=0}^j (-1)^{j-k}\binom{(K-1)-k}{j-k} B_k
\end{align*}
\end{proof}

\section{Disconnectedness of 2-D Orthogonal Convolutions}
\label{appendix:2d-disconnectedness}
\twodbcop*
\begin{proof}
Consider a 2-D convolutional kernel $A$:
\begin{align*}
    A = \begin{pmatrix}
        A_1 & A_2 \\
        A_3 & A_4
    \end{pmatrix}.
\end{align*}

The orthogonality constraint implies that
\begin{align*}
    A_1A_4^T &= 0 \\
    A_2A_3^T &= 0 \\
    A_1A_2^T + A_3A_4^T &= 0 \\
    A_1A_3^T + A_2A_4^T &= 0 \\
    \sum_i A_iA_i^T &= I,
\end{align*}
which yields
\begin{align*}
(A_1 + A_2)(A_3 + A_4)^T &= 0 \\
(A_1 + A_3)(A_2 + A_4)^T &= 0 \\
(A_1 + A_2 + A_3 + A_4)(A_1 + A_2 + A_3 + A_4)^T &= I
\end{align*}

It is clear to see that $A_1 + A_2 + A_3 + A_4$ is orthogonal; hence, we can always find an orthogonal matrix $H = (A_1 + A_2 + A_3 + A_4)^T$ such that
\begin{equation*}
    H (A_1 + A_2 + A_3 + A_4) = I
\end{equation*}

We can apply the orthogonal matrix to each $A_i$ with the same set of constraints:
\begin{align*}
    \Tilde{A}_i &= H A_i
\end{align*}
Then, we have 
\begin{align*}
(\Tilde{A}_1 + \Tilde{A}_2)(\Tilde{A}_3 + \Tilde{A}_4)^T &= 0 \\
(\Tilde{A}_1 + \Tilde{A}_3)(\Tilde{A}_2 + \Tilde{A}_4)^T &= 0 \\
\Tilde{A}_1 + \Tilde{A}_2 + \Tilde{A}_3 + \Tilde{A}_4 &= I 
\end{align*}

Let $P = \Tilde{A}_1 + \Tilde{A}_2$, $Q = \Tilde{A}_1 + \Tilde{A}_3$. We then have
\begin{align*}
P(I - P)^T &= 0 \\
Q(I - Q)^T &= 0
\end{align*}
which is equivalent of saying that $P, Q \in \mathbb{P}(n)$. 

We first prove that there always exists an orthogonal convolution with arbitrary symmetric projectors $P$ and $Q$ by carefully choosing $A_i$'s:

Let $P$ and $Q$ be the symmetric projectors and $H$ be the orthogonal matrix in BCOP algorithm for $2\times 2$ orthogonal convolution, then we have
\begin{equation*}
    A = H \square \begin{pmatrix}
        \Tilde{A}_1 & \Tilde{A}_2 \\ \Tilde{A}_3 & \Tilde{A}_4
    \end{pmatrix} = H \square \begin{pmatrix}
        PQ & P(I-Q) \\ (I-P)Q & (I-P)(I-Q)
    \end{pmatrix}
\end{equation*}
with all the conditions above satisfied $P = \Tilde{A}_1 + \Tilde{A}_2$, $Q = \Tilde{A}_1 + \Tilde{A}_3$. Thus, we can always obtain an orthogonal convolution with arbitrary $P$ and $Q$.

Since the space of symmetric projectors is separated by rank, we can conclude that the space of $\Tilde{A}_1 + \Tilde{A}_2$ is disconnected and have $n + 1$ connected components ($n$ is the size of the matrices). We denote the space of special orthogonal convolution that has $\rank(\Tilde{A}_1 + \Tilde{A}_2) = p$, $\mathcal{X}_p$. Due to disconnectedness of symmetric projector, $\mathcal{X}_p \cup \mathcal{X}_{p'}$ is path-disconnected for any $p\neq p'$. Similarly, we can denote the space of orthogonal convolution that has $\rank(\Tilde{A}_1 + \Tilde{A}_3) = q$, $\mathcal{Y}_q$. $\mathcal{Y}_{q}$ has similar disconnectedness condition. We can define the intersection of $\mathcal{X}_p$ and $\mathcal{Y}_q$ as $\mathcal{S}_{p, q} = \mathcal{X}_p \cap \mathcal{Y}_q$ for all $p, q$. Previously, we proved that there exists orthogonal convolution for any $P$ or $Q$, so $\mathcal{S}_{p, q} \neq \varnothing$ for all $p, q$.  From the disconnectedness of each of $\mathcal{X}$'s and $\mathcal{Y}$'s, we can conclude that $\mathcal{S}_{p, q}$ is disconnected from $\mathcal{S}_{p', q'}$ for any $p, q, p', q'$ if $(p, q) \neq (p', q')$. 

Up to this point, we have identified $(n+1)^2$ disjoint components of $\begin{pmatrix} \Tilde{A}_1 & \Tilde{A}_2 \\ \Tilde{A}_3 & \Tilde{A}_4 \end{pmatrix}$ (the number is induced by the combintorial selection of $p$ and $q$ from $\{0, 1, \cdots, n\}$). Combining this result with the two connected components in orthogonal matrix $H$, we can conclude that the 2-D $2\times 2$ orthogonal convolution has at least $2(n+1)^2$ connected components.
\end{proof}

\section{Doubling the Channel Size Addresses BCOP Disconnectedness Issues}
\label{sec:double}
\bcopdouble*

\begin{proof}\textbf{(Theorem \ref{thm:bcop-aux})}
We will start by defining two functions $f,f'$ to be "effectively equivalent" if $f'(\mathbf{x})_{1:n} = f(\mathbf{x}_{1:n})$. Notice that the input and output are image tensors and $1:n$ is across the channel dimension. It is clear to see that if $f,f'$ are effectively equivalent and so are $g,g'$, the $g\circ f$ is effectively equivalent to $g' \circ f'$. The same holds for $f+g$ and $f'+g'$.

Now we will construct the parameters of $C'$ from the parameters of $C$. For any $n\times n$ projector $P$ used as a parameter for $C$, define $P'$ as,
\begin{align*}
    P' = \begin{pmatrix}
        P & 0 \\
        0 & S_k
    \end{pmatrix}
\end{align*}
where $k$ is the rank of $P$ and $S_k$ is a diagonal matrix with the first $n-k$ entries as 1, and the rest as 0. Notice that $\rank(P') = \rank(P) + \rank(S_k) = k + (n-k) = n$, which independent of the rank of $P$. Finally, replace the orthogonal parameter $H$ for $C$ with 
\begin{align*}
    H' = \begin{pmatrix}
        H & 0 \\
        0 & I
    \end{pmatrix}
\end{align*}
This covers all the parameters in BCOP, so we can now construct $C'$ from $C$ using only $n$-rank projectors. It is easy to see that each projector $P'$ and $H'$ were constructed such that they are effectively equivalent to their counterpart in $C$. Now notice the convolution $C'$ is computed as
\begin{align*}
    C' = H' \square \begin{bmatrix}P'_1 & I-P'_1\end{bmatrix} \square \begin{bmatrix}Q'_1 \\ I-Q'_1\end{bmatrix} \square \cdots \square \begin{bmatrix} P'_{K-1} & I-P'_{K-1}\end{bmatrix} \square \begin{bmatrix} Q'_{K-1} \\ I-Q'_{K-1}\end{bmatrix}
\end{align*}
and by the properties of block convolution, this equivalent to applying each of these functions from right to left to an input image tensor. Thus, by the composition rule for effective equivalence, if all of these functions are equivalent to their $n$ channel counterpart, then $C'$ is effectively equivalent to $C$. Each of these size 2 convolutions computes each output position as a function of two input positions, By first applying a projector to each ($P'$ is applied to one and $I-P'$ is applied to another) and then summing. By the properties of effective equivalence, this means that the application of each size 2 convolution is effectively equivalent to its corresponding convolution in $C$. Multiplying by the orthogonal matrix $H'$ is also effectively equivalent to $H$. Therefore, each function comprising $C'$ is effectively equivalent to its counterpart in $C$, so $C'$ is is effectively equivalent to $C$.
\end{proof}

The theorem above shows that we can represent any $n$-dimensional BCOP convolution with a $2n$-dimensional BCOP convolution that only has rank $n$ projectors, which is a connected space. This allows us to circumvent the disconnectedness issues that arise in our analysis in Appendix \ref{appendix:2d-disconnectedness} by doubling the size of all symmetric projectors by a factor of 2, and set them to be exactly half of the full rank.

However, in order to use this, the input would need $n$ ``dummy'' dimensions so that it can pass through a $2n$-dimensional convolution. This can simply be set up in the initial convolution of a BCOP-parameterized network. The initial convolution would initially have an orthogonal $n\times m$ matrix $H$ to upsample from the network input's $m$ channel size to the desired $n$ channel size ($m$ is typically significantly less than $n$). If we expand $H$ by simply adding $n$ rows of zeros underneath, then we will have a $2n\times m$ matrix which preserves the first $n$ channels of the output while also maintaining orthogonality of $H$. The projectors in this upsampling layer are $n\times n$, so we can simply enlarge these in the same way as in the above theorem to get the desired result from this layer. Finally, we need to address the transition from the convolution layers to the fully-connected layer. This is also straightforward as we can add $n$ columns of 0s to the first fully-connected layer's weight matrix. In this way, the dummy dimensions will have no impact on the network's output. Thus, any network using BCOP convolutions can be equivalently represented in a single connected component of networks using BCOP convolutions with double the channel size. Therefore, this presents a way to circumvent the disconnectedness issue.

\section{Incompleteness of 2-D Convolution Parameterization}
\label{appendix:2d-incompleteness}
\citet{xiao2018dynamical} extends the 1-D orthogonal convolution construction algorithm presented in \citet{kautsky1994matrix} to construct 2-D orthogonal kernel as follows:
\begin{equation}
\begin{aligned}
    A = H \square& \begin{bmatrix} P_1 \\ I-P_1 \end{bmatrix} \square \begin{bmatrix}Q_1 & I-Q_1 \end{bmatrix} \square \cdots \\ \cdots&\square \begin{bmatrix} P_{K-1} \\ I-P_{K-1} \end{bmatrix}\square \begin{bmatrix} Q_{K-1} & I-Q_{K-1} \end{bmatrix}
\end{aligned}
\end{equation}
where $X \square Y = \left\lbrack \sum_{i', j'} X_{i', j'} Y_{i - i', j - j'} \right\rbrack_{i, j}$ with the out-of-range matrices all zero, $H$ is an orthogonal matrix, and $P_1, \cdots P_{K-1}$ and $Q_1, \cdots Q_{K-1}$ are symmetric projectors. However, unlike in the 1-D case, the 2-D orthogonal kernels constructed from this algorithm doesn't cover the entire orthogonal kernel space.

To illustrate this, we first consider a $2\times2$ convolution kernel $A = \begin{bmatrix} A_1 & A_2 \\ A_3 & A_4 \end{bmatrix}$ with its equivalent transformation matrix over a $3 \times 3$:
\begin{equation*}
    \begin{bmatrix}
        A_1 & A_2 & 0 & A_3 & A_4 & 0 & 0 & 0 & 0 \\
        0 & A_1 & A_2 & 0 & A_3 & A_4 & 0 & 0 & 0 \\
        A_2 & 0 & A_1 & A_4 & 0 & A_3 & 0 & 0 & 0 \\
        0 & 0 & 0 & A_1 & A_2 & 0 & A_3 & A_4 & 0 \\
        0 & 0 & 0 & 0 & A_1 & A_2 & 0 & A_3 & A_4 \\
        0 & 0 & 0 & A_2 & 0 & A_1 & A_4 & 0 & A_3 \\
        A_3 & A_4 & 0 & 0 & 0 & 0 & A_1 & A_2 & 0 \\
        0 & A_3 & A_4 & 0 & 0 & 0 & 0 & A_1 & A_2 \\
        A_4 & 0 & A_3 & 0 & 0 & 0 & A_2 & 0 & A_1 \\
    \end{bmatrix}
\end{equation*}

The equivalent conditions for the convolution kernel to be orthogonal can be summarized as follows (inner products of any pairs of distinct rows need to be 0):
\begin{equation*}
\begin{aligned}
    A_1 A_4^T &= 0 \text{  (Inner product of Row 1 and Row 9)}\\
    A_2 A_3^T &= 0 \text{  (Inner product of Row 1 and Row 8)}\\
    A_1A_2^T + A_3A_4^T &= 0 \text{  (Inner product of Row 1 and Row 2)}\\
    A_1A_3^T + A_2A_4^T &= 0 \text{  (Inner product of Row 1 and Row 7)}\\
    \sum_i A_i A_i^T &= I \text{  (Self inner product of any row)} 
\label{eq:2x2-orthogonal-constraint}
\end{aligned}
\end{equation*}

However, notice that since BCOP $2\times 2$ convolution is of the form,
\begin{equation*}
\begin{aligned}
    A = H \square& \begin{bmatrix} P \\ I-P \end{bmatrix} \square \begin{bmatrix}Q & I-Q \end{bmatrix} = \square &\begin{bmatrix}HPQ & HP(I-Q) \\ H(I-P)Q & H(I-P)(I-Q)\end{bmatrix}
\end{aligned}
\end{equation*}
It is clear we will always have an additional constraint of $A_1A_2^T = (HPQ)(HP(I-Q))^T = 0$. However, we can find an orthogonal matrix that does not satisfy this condition by defining the $A_1$ to $A_4$ as:

\begin{equation*}
\begin{aligned}
    A_1 = \frac{1}{2}\begin{bmatrix}1 & 0 \\ -1 & 0\end{bmatrix} \\
    A_2 = \frac{1}{2}\begin{bmatrix}1 & 0 \\ 1 & 0\end{bmatrix} \\
    A_3 = \frac{1}{2}\begin{bmatrix} 0 & -1 \\ 0 & 1\end{bmatrix} \\
    A_4 = \frac{1}{2}\begin{bmatrix}0 & 1 \\ 0 & 1\end{bmatrix}
\end{aligned}
\end{equation*}

However, while this does show that BCOP is incomplete, the counterexample is fairly uninteresting. In fact, it can be represented as
\begin{align*}
    A = \begin{bmatrix} P & I-P \end{bmatrix} \square \begin{bmatrix}Q \\ I-Q \end{bmatrix},\ \text{where } P = \begin{bmatrix} 1 & 0 \\ 0 & 0 \end{bmatrix},\ Q = \frac{1}{2}\begin{bmatrix} 1 & -1 \\ -1 & 1 \end{bmatrix}
\end{align*}
which is simply a re-ordering of the 1-D convolutions in the BCOP definition. Furthermore, it can even be trivially represented by composing two BCOP convolutions together. It is still an open question as to whether or not allowing re-ordering of the 1-D convolution components of BCOP will be able to represent all 2-D orthogonal convolutions (while re-ordering 1-D convolutions is not something easily achievable in neural network architectures, composing arbitrarily many BCOP convolutions can represent any re-ordering of 1-D convolutions).

\section{Certifying Provable Robustness for Lipschitz Network}
\label{sec:appendix-adv-robust}
To certify provable robustness of a Lipschitz network, we first define the margin of a prediction for a data point $x$,
\begin{equation*}
    \mathcal{M}_f(\mathbf{x}) = \max(0, y_t - \max_{i\neq t} y_i)
\end{equation*}
where $\mathbf{y} = [y_1, y_2, \cdots]$ is the predicted logits from the model on data point $\mathbf{x}$ and $y_t$ is the correct logit ($\mathbf{x}$ belongs to $t^{\text{th}}$ class).
Following the results from \citet{anil2018sorting, tsuzuku2018lipschitz}, we derive the sufficient condition for a data point to be provably robust to perturbation-based adversarial examples in the general case: 

\begin{theorem}[Adversarial Perturbation Robustness Condition under $L_p$ Norm]
If $2^{\frac{p-1}{p}}l\epsilon < \mathcal{M}_f(\mathbf{x})$, where $f$ is an $l$-Lipschitz under the $L_p$ norm, then $\mathbf{x}$ is robust to any input perturbation $\Delta \mathbf{x}$ with $||\Delta \mathbf{x}||_p \leq \epsilon$. 
\label{theorem:adv-robust-condition-lp}
\end{theorem}

\begin{proof}\textbf{(Theorem \ref{theorem:adv-robust-condition-lp})}\\
Let the function that represents the network to be $f$, $\mathbf{x}$ be some data point, and $\mathbf{y} = f(\mathbf{x})$.

We see that it is enough to consider when $\mathbf{x}$ is correctly classified. If it were misclassified, $\mathcal{M}_f(\bx)=0$, so $2^{\frac{p-1}{p}}l\epsilon < \mathcal{M}_f(\bx)$ can never hold. Since $\mathbf{x}$ is correctly classified, $\mathcal{M}_f(\bx) \leq y_t - y_i$ for all $i \neq t$ where $t$ is the correct class. 

Now suppose there is a $\Delta \mathbf{x}$ such that $||\Delta \mathbf{x}||_p \leq \epsilon$ and $\mathbf{x}' = \mathbf{x} + \Delta \mathbf{x}$ is incorrectly classified. Then, $y_t' \leq y_w'$ for some $w$ where $\mathbf{y'}=f(\mathbf{x}')$. We also denote the perturbation in the $i^\text{th}$ dimension to be $\Delta y_i = y_i'-y_i$.

Since $f$ is $L$-Lipschitz, we can bound the norm of the change of output as follows:
\begin{equation*}
      \left({\sum_i |\Delta y_i|^p}\right)^{\frac{1}{p}} = ||f(\mathbf{x} + \Delta\mathbf{x}) - f(\mathbf{x})||_p \leq l||\Delta\mathbf{x}||_p \leq l\epsilon
\end{equation*}

Then we have
\begin{equation*}
    |\Delta y_t|^p + |\Delta y_w|^p \leq \sum_i |\Delta y_i|^p \leq (l\epsilon)^p
\end{equation*}

Now consider a polynomial $g(r) = r^p + (\Delta y_w - \Delta y_t - r)^p$. By analyzing derivatives, this has a global minimum at $r = \frac{\Delta y_w - \Delta y_t}{2}$ using the fact that $\Delta y_w \geq \Delta y_t$. This yields,
\begin{equation*}
    \frac{|\Delta y_w - \Delta y_t|^p}{2^{p-1}} = g\left(\frac{\Delta y_w - \Delta y_t}{2}\right) \leq g(-\Delta y_t) = (-\Delta y_t)^p + (\Delta y_w)^p \leq |\Delta y_w|^p + |\Delta y_t|^p \leq (l\epsilon)^p
\end{equation*}

This means that
\begin{equation*}
    |\Delta y_w - \Delta y_t| \leq 2^{\frac{p-1}{p}}l\epsilon \Rightarrow -2^{\frac{p-1}{p}}l\epsilon \leq \Delta y_t - \Delta y_w
\end{equation*}

Substituting this bound along with the inequality $y_t - y_w \geq \mathcal{M}_f(\mathbf{x})$ yields
\begin{equation*}
    0 \geq y'_t - y'_w = y_t - y_w + (\Delta y_t - \Delta y_w) \geq \mathcal{M}_f(\mathbf{x}) - 2^{\frac{p-1}{p}}(l\epsilon).
\end{equation*}
Therefore, $\mathcal{M}_f(\mathbf{x}) \leq 2^{\frac{p-1}{p}}(l\epsilon)$. 

Thus, by contrapositive $\mathcal{M}_f(\mathbf{x}) > 2^{\frac{p-1}{p}}(L\epsilon)$ implies $y'_t - y'_w > 0$.
\end{proof}

\end{document}